\documentclass{article}

\usepackage[main, final]{neurips_2026}

\usepackage[utf8]{inputenc}
\usepackage[T1]{fontenc}
\usepackage{microtype}

\usepackage{amsmath}
\usepackage{amssymb}

\usepackage{graphicx}
\usepackage{booktabs}
\usepackage{enumitem}
\usepackage[svgnames]{xcolor}

\usepackage{hyperref}
\usepackage[capitalize]{cleveref}
\hypersetup{
    colorlinks=true,
    allcolors=DarkBlue,
    pdftitle={Measuring Coherence in Predictive Models},
    pdfauthor={Jamie Reason, Vik Shirvaikar, Mark van der Wilk, Chris Holmes},
}

\DeclareMathOperator{\MMD}{MMD}
\DeclareMathOperator{\TV}{TV}
\DeclareMathOperator{\KL}{KL}
\DeclareMathOperator{\EPIG}{EPIG}
\DeclareMathOperator{\EMI}{EMI}
\DeclareMathOperator{\Var}{Var}
\DeclareMathOperator{\clip}{clip}
\DeclareMathOperator{\softmax}{softmax}
\DeclareMathOperator{\round}{round}

\title{Measuring Coherence in Predictive Models}

\author{%
    Jamie Reason\thanks{Correspondence: \texttt{jamie.reason@reuben.ox.ac.uk}. Code: \href{https://github.com/jreaso/predictive-coherence}{\texttt{github.com/jreaso/predictive-coherence}}.} \\
    University of Oxford
    \And
    Vik Shirvaikar \\
    University of Oxford
    \And
    Mark van der Wilk \\
    University of Oxford
    \And
    Chris Holmes \\
    University of Oxford
}

\begin{document}

\maketitle

\begin{abstract}
Many sequential decision-making procedures, such as active learning, implicitly assume that a model's current predictions agree with those it expects to make after updating on new data, a property we call coherence.
Most models in use do not update by exact conditioning and are not guaranteed to be coherent, which can lead to worse decisions.
Yet there is currently no general way to measure incoherence or to tell when it matters.
We develop tools to measure incoherence and characterise when it affects decisions.
We introduce the incoherence ratio, an MMD-based measure of how far any updating procedure departs from coherence, computed only from the procedure's own predictions and updates, and accompany it with an anytime-valid test of coherence.
In Bayesian experimental design, acquisition objectives that are equivalent for coherent models choose different data under every approximate model we study, and the incoherence ratio tracks how far incoherence distorts them.
We also introduce action coherence, which captures only the incoherence that changes a decision.
What a resulting disagreement costs, however, is determined primarily by predictive accuracy.
Sequential decisions depend on how a model updates as well as on what it predicts, so coherence, like accuracy, should be checked rather than assumed.
\end{abstract}

\section{Introduction}
\label{sec:intro}

In sequential decision-making problems, actions are often chosen using a model that is updated as new data arrive.
Active learning~\citep{mackay1992information, cohn1996active}, for example, selects the data whose acquisition will most improve the model's predictions after updating.
Making this choice requires the model to anticipate how its own predictions will change, which relies on its current predictions agreeing with those it expects to make after updating.
Following \citet{fong2023martingale}, we call such a model coherent.
Many methods for sequential decision-making are derived assuming exact Bayesian conditioning~\citep{rainforth2024bed, garnett2023bayesian}, which guarantees coherence by construction.
The models they are applied to, however, rarely update by exact conditioning, and so need not be coherent.
Rather than assuming coherence, we measure how severely it is violated, and identify when and how incoherence affects decisions.

Bayesian experimental design shows how incoherence can reach a decision.
The expected predictive information gain (EPIG)~\citep{bickfordsmith2023prediction} selects the design whose outcome most reduces uncertainty in the model's prediction at a target of interest.
It can be written in several standard forms: a reduction in predictive entropy, a mutual information, or an expected divergence from the current predictive, and these are used interchangeably because they agree under coherence.
For an incoherent model, however, they become different acquisition rules that can rank candidate designs in opposite orders (\Cref{sec:cost}).
Empirically, the entropy form acquires worse data on average than the mutual-information form for every approximate model we study (\Cref{sec:experiments}).

Coherence has a long history in both philosophy~\citep{vanfraassen1984belief, lewis1999conditionalize} and statistics~\citep{goldstein1983prevision}, where it corresponds to the martingale property of predictive distributions~\mbox{\citep{fong2023martingale}} and underpins the predictive approach to Bayesian inference~\citep{definetti1937prevision}.
Most models central to machine learning do not update by explicit conditioning, and a growing body of work asks whether their updates nonetheless behave like a Bayesian update.
Prediction rules have been measured against exchangeability~\citep{mlodozeniec2024implicitly}, in-context learning in large language models and tabular foundation models has been shown to depart from the martingale property~\citep{falck2024incontext, ng2026tabmgp}, and approximate Bayesian neural networks have been found to update incoherently~\citep{pituk2025bayesian}.
These works establish that coherence can fail in important models, using diagnostics specific to each setting, most of which need data from the task or a reference Bayesian model.
Separately, work on sequential decisions has shown that decision quality depends on a model's joint predictive distribution, rather than its marginal predictions~\citep{wen2022from, osband2022neural}.
A model's update rule implicitly defines such a joint, and coherence is the requirement that it agree with the model's current predictions.
Because coherence concerns how predictions update rather than how accurate they are, a model can predict accurately at every input and still be incoherent.
What is still missing is a general way to measure the incoherence of an arbitrary updating procedure using only the procedure itself, and an account of when incoherence matters for the decisions these models are used to make.

We address both.
We first set out what coherence means for any procedure that predicts and updates, without reference to a prior, likelihood or parameter posterior (\Cref{sec:coherence}).
The requirement is local to the history, target and update relevant to the problem, and, in keeping with a predictivist view, concerns only predictions of observable quantities.
To measure incoherence, we introduce the incoherence ratio \(\Psi\), a scale-invariant measure of how far a procedure's updates depart from conditioning, estimated from its own predictions and updates alone (\Cref{sec:measuring-incoherence}).
We also give an anytime-valid sequential test of coherence (\Cref{sec:testing-coherence}).
To determine when incoherence matters, we introduce action coherence, a relaxation that asks only whether incoherence changes the action a procedure selects, and so isolates exactly what reaches a decision (\Cref{sec:cost}).

Our experiments show what incoherence reveals about models and when it reaches decisions (\Cref{sec:experiments}).
Our measure of incoherence falls as an approximation is refined and ranks models consistently across datasets.
For every approximate model we study, objectives that agree under coherence select different data on a substantial share of decisions, so incoherence reaches the action.
The measure tracks how susceptible a model's decisions are to this disagreement, while accuracy determines what it costs.
Incoherence is therefore a warning rather than a verdict.
Sequential decisions rely on how a model updates, not only on what it predicts, and coherence is the property that captures this.

\section{Predictive coherence}
\label{sec:coherence}

A predictive procedure comprises a probabilistic model and its prescribed update rule.
At a realised history \(h\), the procedure returns predictive laws for the variables required by the prediction and update problem.
We write \(Y\) for the prediction target and \(Z\) for the future observations that will be incorporated before the procedure is queried for \(Y\) again.
We call \(Z\) the \emph{update block}, which may contain one or more observations.
An ordered update design \(d=(d_1,\ldots,d_K)\) specifies which observations comprise the block and their order.
The update is written \(\mathsf{U}(h,z)\) and maps a history and an observed update block to an updated history.
In supervised learning, for example, the initial history may be the observed dataset \(h=(x_{1:n},y_{1:n})\), and the target \(Y\) the response at a fixed test input \(x_*\).
An update design \(d = x_{n+1:n+K}\) specifies a collection of additional inputs, with corresponding update block \(Z = Y_{n+1:n+K}\).
Here the components of \(Z\) are responses of the same kind as \(Y\).
More generally, however, the construction assumes neither that the observations are exchangeable nor that the components of \(Z\) take values in the same space as \(Y\).

We write \(P_Y(\cdot;h)\) and \(P_Z(\cdot;h)\) for the procedure's current predictive laws for the target and the complete update block, respectively.
When the procedure does not supply a joint law for \(Z=(Z_1,\ldots,Z_K)\), we construct one prequentially~\citep{dawid1984present} in the order specified by the design \(d\), as in predictive resampling~\citep{fong2023martingale}: \(P_Z(dz ; h) =  \prod_{k=1}^K P_{Z_k}\!\left(dz_k ; h_{k-1} \right)\), where \(h_0 \equiv h\) and \(h_k := \mathsf{U}(h_{k-1}, z_k)\) is the updated history after observing \(z_{\leq k}\).
We assume throughout that updating on the whole block agrees with updating component-wise along \(d\), so that \(h_K = \mathsf{U}(h,z)\).
Throughout, a semicolon indicates that a predictive law is indexed by a history, and we reserve the vertical bar for probabilistic conditioning within a specified joint distribution.
Subscripts name the variable whose law is described and superscripts carry the indices on which the law depends, namely the update design and its realisation.
We suppress any index that is held fixed in the surrounding discussion, and reinstate it where it varies.
In this section the design is fixed and suppressed.

\paragraph{Update-induced joint.}
The model's predictive law for the update block and its post-update target predictions together define the \emph{update-induced joint},
\begin{equation}
    Q(dy,dz;h) := P_Y\!\left(dy;\mathsf U(h,z)\right)P_Z(dz;h).
    \label{eq:update-induced-joint}
\end{equation}
By construction, \(Q_Z(\cdot;h)=P_Z(\cdot;h)\) and \(Q_{Y\mid Z=z}(\cdot;h) = P_Y\!\left(\cdot;\mathsf{U}(h,z)\right)\) for \(P_Z(\cdot;h)\)-almost every \(z\).
Thus \(Q\) is the unique joint distribution whose \(Z\)-marginal represents the model's current beliefs about the update block and whose conditional distributions represent the predictions that the procedure will actually return after updating.
The target marginal of this joint is
\begin{equation}
    Q_Y(dy;h) = \int P_Y\!\left(dy;\mathsf{U}(h,z)\right)P_Z(dz;h).
    \label{eq:expected-future-predictive}
\end{equation}
This is the model's post-update predictive averaged under its current beliefs about the update block, which we call the \emph{expected future predictive}.
The update-induced joint is the operationally relevant distribution for a procedure which first observes \(Z\) and then \(Y\).
When the history \(h\) is fixed, we suppress it, writing \(P_Y\) and \(P_Z\) for the current predictives, \(P_Y^z := P_Y(\cdot; \mathsf{U}(h,z))\) for the post-update predictive, and \(Q_Y\) for the expected future predictive, so that \(Q_Y = \mathbb{E}_{P_Z}[P_Y^Z]\) and \(Q_{Y\mid z} \equiv P_Y^z\).

\paragraph{Predictive coherence.}
A predictive procedure is coherent for the target \(Y\), history \(h\) and update block \(Z\) when the current predictive and expected future predictive agree, \(P_Y=Q_Y\).
It extends the martingale condition of \citet{fong2023martingale} to any target, history and update block.
Coherence is exactly the statement that the current predictive is recovered as the \(Y\)-marginal of the update-induced joint.
If coherence fails, no joint can simultaneously have \(P_Z\) as its \(Z\)-marginal, \(P_Y^z\) as its conditional distribution given \(Z=z\), and \(P_Y\) as its \(Y\)-marginal.
Exact conditioning under a model-specified joint is sufficient for predictive coherence.
The general definition, however, does not require the model to specify a joint law over \((Y,Z)\) independently of the update-induced joint \(Q\).

\paragraph{Local coherence.}
Predictive coherence is a local condition: it is defined relative to a realised history \(h\), a target \(Y\), and a specified update setup comprising the ordered design \(d\) and its corresponding update block \(Z\).
Coherence for one such setup does not imply coherence at another history, for another target, or under another update design.
This locality is both a limitation and a feature.
It limits the scope of any coherence guarantee, but focuses the requirement on the predictions and updates that are relevant to the problem under consideration.
Incoherence outside this scope need not affect the intended use of the procedure.
From a predictivist perspective, coherence concerns the compatibility of predictions about the observable quantities \(Y\) and \(Z\) across the histories reachable within this update setup.
It therefore makes no general claim about the model's latent parameters, posterior approximation, or global Bayesianity.
We call a predictive procedure globally coherent only if the coherence condition holds for every admissible history, target, and update design.

\paragraph{Incoherence admits a Dutch book.}
Incoherence can have direct consequences for tasks that rely on a predictive procedure.
The canonical demonstration is a \emph{diachronic Dutch book}: a sequence of contracts that the procedure regards as individually fair at the histories where they are offered, but whose combined payoff is a sure loss.
The classical construction of \citet{lewis1999conditionalize} shows qualitatively that such exploitation is possible whenever the procedure's current valuation of some payoff differs from its expected valuation after updating.
We quantify the maximum severity of this exploitation under the betting protocol and a bounded-liability constraint.
Writing \(D_{\TV} := \TV(P_Y, Q_Y)\), the maximum fraction of initial wealth which can be extracted as a sure loss is \(D_{\TV} / (1 + D_{\TV})\).
This Dutch-book result concerns the internal consistency of the procedure's valuations and it does not imply that either predictive distribution is empirically inaccurate.
\Cref{app:dutch-book} gives the classical construction and derives the maximum sure-loss fraction.

\subsection{Bayesian models and coherence}

Exact Bayesian conditioning provides the canonical sufficient condition for predictive coherence, but it is often unavailable.
The familiar prior-likelihood construction in Bayesian deep learning has an intractable parametric posterior which must be approximated.
Practical approximations include sampling-based methods such as SG-MCMC~\citep{ma2015complete}, variational inference over a restricted family~\citep{blei2017variational}, such as the diagonal Gaussian used by IVON~\citep{shen2024variational}, and the linearised Laplace approximation~\citep{immer2021improving}.
Whilst each targets a coherent Bayesian model, the approximations can induce incoherence in the resulting predictive procedure.

Amortised approaches such as neural processes~\citep{garnelo2018neural, garnelo2018conditional} and prior-data fitted networks~\citep{muller2022transformers} bypass explicit parametric inference and learn a direct map from an observed dataset to a predictive distribution.
They approximate the posterior predictive defined by their training prior, and in practice they too fall short of exact coherence, as observed for TabPFN~\citep{ng2026tabmgp}.
Similarly, deep ensembles~\citep{lakshminarayanan2017simple} can be interpreted as approximating Bayesian model averaging~\citep{wilson2020bayesian}, despite their members not generally being samples from the posterior of a specified BNN~\citep{wild2023rigorous}.
These constructions demonstrate that a procedure can be approximately Bayesian without an explicit target model.
A procedure's Bayesian provenance does not determine how severe its incoherence will be, and we therefore separate construction from empirical behaviour.

\section{Cost of incoherence}
\label{sec:cost}

Incoherence matters when an action is taken before an update but valued by the predictions the procedure will report after it.
We consider a decision problem spanning a single update.
An initial action \(a \in \mathcal{A}\) is taken, a block \(Z^a\) is observed, the procedure is updated, a terminal action \(b \in \mathcal{B}\) is chosen from the post-update predictive, and a loss \(\ell(a,b,y)\) is realised at the target.
The design of the block may depend on the initial action, so we index by the action, writing the update-induced joint as \(Q^a(dy,dz) = P_Y^{a,z}(dy) P_Z^a(dz)\).
Prediction-oriented experimental design is a problem of this form, and problems spanning several updates, as in Bayesian optimisation~\citep{garnett2023bayesian}, are covered when each action is chosen greedily.
The terminal action depends on the procedure only through the post-update predictive, which is a conditional of \(Q^a\).
Of the predictions consumed in valuing the initial action, only the current predictive \(P_Y\) can therefore disagree with the update-induced joint \(Q^a\).

\paragraph{Action coherence.}
We \emph{coherentise} an objective by substituting \(Q_Y^a\) for \(P_Y\) wherever it appears, writing \(J\) for the stated objective and \(J^+\) for its coherentisation, and \(a^*\) and \(a^+\) for the actions they select.
A procedure is \emph{action coherent} at \(h\) for \((\mathcal{A}, J)\) when coherentising leaves the decision unchanged, \(a^* = a^+\), with ties broken identically.
Predictive coherence for every block in \(\{Z^a : a \in \mathcal{A}\}\) implies action coherence for every objective, but not conversely, so action coherence is a relaxation that registers only the incoherence that reaches the action.
Objectives that coincide under coherence share the same coherentisation, so two such objectives selecting different actions is sufficient for at least one of them to be action incoherent.
Coherentisation is internal to the procedure and is not a correction towards the truth, since \(Q^a\) may describe the true process poorly.
It removes inconsistency, not inaccuracy.

\paragraph{Severity of incoherence.}
What reaches the action is not the size of the discrepancy between \(Q_Y^a\) and \(P_Y\), but how it varies across candidates relative to how sharply the objective separates them, and how the objective consumes \(P_Y\).
A large discrepancy that is near-constant across candidates barely moves the decision, so detecting predictive incoherence establishes that a decision may be affected without indicating by how much.
Incoherence also decouples objectives that coincide under coherence, and no member of the decoupled family is the correct one, since the current predictive and the expected future predictive disagree and neither is guaranteed to be the more accurate.
Which member selects better actions under the true law is an empirical question we address in \Cref{sec:experiments}.

\subsection{Objective splitting in experimental design}
\label{sec:objective-splitting}

Bayesian experimental design (BED)~\citep{rainforth2024bed} selects designs which maximise information about a downstream quantity.
In prediction-oriented design, that quantity is the procedure's prediction at deployment.
Here \(\mathcal{A}\) is a set of candidate designs, \(Z^a\) is the observation acquired at the chosen design, and the terminal action reports a predictive for a target \(Y\) under the log score.
The expected predictive information gain (EPIG)~\citep{bickfordsmith2023prediction} acquires the design which most reduces predictive uncertainty at the target, \(\EPIG_H(a) = H[P_Y] - \mathbb{E}_{Z \sim P_Z^a} [H[P_Y^{a,Z}]]\), which with the procedure's own update is the expected uncertainty reduction of \citet{bickfordsmith2025rethinking}.
Read decision-theoretically, EPIG is a design-free constant minus the procedure's expected log loss after the update, so maximising it minimises that loss.
Read information-theoretically, it is the mutual information between the target and the acquired observation under the joint the procedure induces over them.
Under incoherence that joint has \(Y\)-marginal \(Q_Y^a\) rather than \(P_Y\), and the mutual information becomes \(\EMI(a) = H[Q_Y^a] - \mathbb{E}_{Z \sim P_Z^a} [H[P_Y^{a,Z}]]\), the coherentisation of \(\EPIG_H\).
The two differ only in their baseline, which is constant in \(a\) for \(\EPIG_H\) but varies with \(a\) once coherentised, and so can change the ranking.
A third standard form is the expected divergence from the current predictive, and both it and \(\EPIG_H\) depart from EMI by explicit terms (\Cref{app:epig-forms}),
\begin{align}
    \EPIG_{\KL}(a) := \mathbb{E}_{Z \sim P_Z^a}\!\left[\KL\!\left(P_Y^{a,Z} \,\|\, P_Y\right)\right] &= \EMI(a) + \KL\!\left(Q_Y^a \,\|\, P_Y\right), \label{eq:epig-kl-identity}\\
    \EPIG_H(a) &= \EMI(a) + H[P_Y] - H[Q_Y^a]. \label{eq:epig-h-identity}
\end{align}
All three forms coincide under coherence and coherentise to EMI, but under incoherence they are different acquisition rules, so even a choice ordinarily made for numerical convenience becomes a choice of acquisition rule.
The KL form departs from EMI by a non-negative term that is second order in the discrepancy between \(Q_Y^a\) and \(P_Y\), approximately half their \(\chi^2\) divergence, whereas the entropy form departs by a signed term that is first order.
For small incoherence the KL form therefore stays much closer to EMI, as \Cref{sec:experiments} confirms.

\begin{figure}[t]
    \centering
    \includegraphics[width=0.99\textwidth]{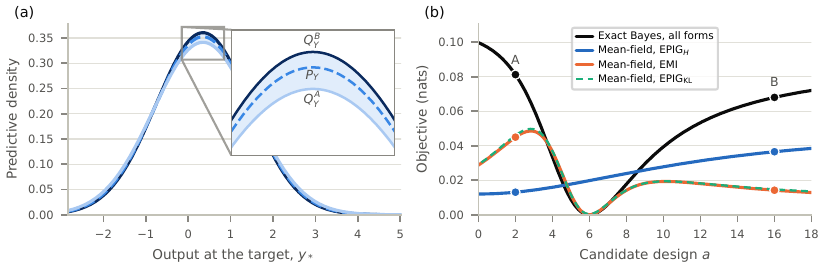}
    \caption{Objective splitting under mean-field VI for Bayesian linear regression.
    \textbf{(a)}~The mean-field model's current predictive \(P_Y\) (dashed) and its expected future predictives \(Q^A_Y\) and \(Q^B_Y\) after acquiring candidates \(A\) and \(B\).
    The inset magnifies the peak, where neither expected future predictive matches the current one, so the model is incoherent for both candidates.
    \textbf{(b)}~The three forms of EPIG against the candidate design \(a\).
    Under exact Bayes they coincide.
    Under mean-field VI, \(\EPIG_H\) prefers \(B\) whereas EMI prefers \(A\), and \(\EPIG_{\KL}\) (dashed) stays within 0.002 nats of EMI.}
    \label{fig:bed-objective-splitting}
\end{figure}

We make the split concrete with two-parameter Bayesian linear regression under a Gaussian likelihood and prior, comparing exact conditioning against mean-field variational inference, whose factorised posterior cannot represent the intercept--slope correlation induced by training inputs clustered away from the origin.
All three forms are available in closed form, so any discrepancy is attributable to incoherence rather than estimation noise.
Sweeping the candidate design (\Cref{fig:bed-objective-splitting}), they coincide under exact conditioning, and under mean-field VI \(\EPIG_H\) and EMI rank two candidates in opposite orders.
\Cref{sec:experiments} measures how often this occurs, and what it costs, across model families.

\section{Measuring incoherence}
\label{sec:measuring-incoherence}

\Cref{sec:cost} showed that incoherence can reach decisions.
Since few procedures update by exact conditioning, the question that usually matters in practice is how incoherent a procedure is, rather than whether it is exactly coherent, which we return to in \Cref{sec:testing-coherence}.
This section develops a scalar measure of that severity, computed from the procedure's own predictions and updates.

\subsection{The incoherence ratio}
\label{sec:incoherence-ratio}

Fix a history \(h\), a target \(Y\) and a design \(d\) with block \(Z\).
Each draw from \(Q_Y\) costs an update, so updates are the budget, and we treat the procedure as a black box that can be updated and sampled.
A \emph{round} draws a block \(Z_r \sim P_Z\), updates the fitted procedure on it, drawing any randomness internal to the update afresh, and returns \(P_Y^{Z_r}\), so rounds are independent and identically distributed.
The subscript on \(Z_r\) indexes rounds, not the components of the block.
\citet{pituk2025bayesian} use such rounds to compare test metrics, not predictive laws.
The Dutch book of \Cref{sec:coherence} prices incoherence by \(\TV(P_Y, Q_Y)\), but total variation is non-linear in the mixture \(Q_Y\) and cannot be estimated without bias from \(R\) post-update predictives.
We instead use kernel mean embeddings~\citep{gretton2012kernel}, which are linear in the mixture.
For a kernel \(k\) with feature map \(\phi\) and \(\|\phi(y)\|^2 \le \kappa\), write \(\mu_P\), \(\mu_Q\) and \(\mu_r\) for the embeddings of \(P_Y\), \(Q_Y\) and \(P_Y^{Z_r}\) respectively, and \(\delta_r := \mu_r - \mu_P\) for the displacement of round \(r\).
Then \(\Delta := \mu_Q - \mu_P = \mathbb{E}[\delta_r]\) and \(\MMD^2(P_Y, Q_Y) = \|\Delta\|^2\).
For a characteristic kernel~\citep{sriperumbudur2010hilbert}, \(\Delta = 0\) if and only if the procedure is coherent.
Since \(\MMD \le 2\sqrt{\kappa}\, D_{\TV}\), however, \(\|\Delta\|\) bounds the sure loss only from below, and a small \(\Delta\) certifies only that incoherence is small in the kernel's features (\Cref{app:ratio-properties}).

\paragraph{The incoherence ratio.}
The squared discrepancy \(\|\Delta\|^2\) grows with how far a procedure's updates move its predictive, so procedures that update by different amounts cannot be compared on it.
The bias--variance identity \(\mathbb{E}\|\delta_r\|^2 = \|\Delta\|^2 + V\), with \(V := \mathbb{E}\|\delta_r - \Delta\|^2\), separates the systematic part of an update from its response to the block, and we define the \emph{incoherence ratio} as
\begin{equation}
    \Psi := \frac{\|\Delta\|^2}{V} = \frac{\MMD^2(P_Y, Q_Y)}{\mathbb{E}\bigl[\MMD^2(P_Y^{Z_r}, Q_Y)\bigr]},
    \label{eq:incoherence-ratio}
\end{equation}
the squared signal-to-noise ratio of the update.
It is invariant to rescaling the displacements, and \(\Psi/(1+\Psi)\) is the systematic share of the expected squared displacement, so \(\Psi = 1\) means bias and spread contribute equally.
For a characteristic kernel, \(\Psi\) vanishes exactly under coherence, and a shift that ignores the block, with \(V = 0\) and \(\Delta \neq 0\), has \(\Psi = \infty\) (\Cref{app:ratio-properties}).
\(\Psi\) is a self-consistency check that needs no test set, reference posterior or oracle.
Blocks are drawn from \(P_Z\) rather than from data because coherence concerns the procedure's beliefs about its own future, and measuring on observed blocks would mix inconsistency with inaccuracy.
Like coherence, \(\Psi\) is a property of the procedure for the update problem \((h, Y, d)\), including the size of the block, which should match the update the procedure will face.
Its level depends on the kernel and the design, so it ranks procedures only within a problem.
Since \(V\) includes any randomness internal to the update, a noisier update has a lower \(\Psi\) for the same bias, and \Cref{app:ratio-properties} separates the spread driven by the block.

\paragraph{Estimation.}
\(\Psi\) is estimated from \(R\) rounds, which can be shared with the test of \Cref{sec:testing-coherence}.
U-statistics over rounds estimate \(\|\Delta\|^2\) and \(V\) without bias, reducing with exact embeddings to \(\|\hat\Delta\|^2 - \hat V/R\) and \(\hat V\), and the ratio has an \(O(R^{-1})\) bias and a leave-one-round-out standard error (\Cref{app:ratio-estimation}).
Near coherence the numerator is a degenerate U-statistic and estimates can be negative, which we report unclipped.
Displacements at several evaluation points are pooled as a ratio of summed numerators to summed denominators, which keeps scale invariance.
For finite targets we use the Dirac kernel, and for continuous targets random Fourier features~\citep{rahimi2007random}, which make estimation cheap but are not characteristic.
We set their lengthscale to the noise scale of the problem, and in \Cref{sec:experiments}, \(\Psi\) is found to be robust to this choice.

\subsection{Measuring action incoherence}
\label{sec:measuring-action-incoherence}

\(\Psi\) is decision-agnostic.
Whether incoherence reaches an action is defined by the coherentised objective \(J^+\), which is rarely computable, since it consumes \(Q_Y^a\) through a non-linear functional at every candidate.
Because objectives that coincide under coherence share a coherentisation (\Cref{sec:cost}), the rate at which two tractable objectives from this class disagree witnesses action incoherence, although agreement does not certify its absence.
The two measures answer different questions.
The disagreement rate concerns a particular pair of objectives and decision problem.
\(\Psi\) needs neither, so it can compare procedures before a decision problem is fixed, audit an approximation choice, and attribute a failed sequential procedure to inconsistent updates rather than inaccurate predictions.
The link runs in one direction only.
With a characteristic kernel, if \(\Psi = 0\) for every block in \(\{Z^a : a \in \mathcal{A}\}\), then the procedure is action coherent for every objective.
A positive \(\Psi\) diagnoses susceptibility, the distortion of the objective across candidates.
Whether that distortion changes a choice also depends on the margin between candidates, and what a changed choice costs is evaluated under the true law rather than the procedure's own beliefs (\Cref{sec:cost}), so neither is determined by \(\Psi\).
Coherentisation removes inconsistency rather than inaccuracy, so realised performance on an oracle or held-out set remains the decision-relevant comparison, and \Cref{sec:experiments} examines how susceptibility relates to it.

\section{Testing coherence}
\label{sec:testing-coherence}

The incoherence ratio grades how incoherent a procedure is, but its estimate is not a test of coherence.
Near coherence its numerator is a degenerate U-statistic, for which valid intervals need a dedicated construction~\citep{manole2023martingale}, and \(\hat\Psi \pm 2\,\mathrm{SE}\) is not a confidence interval.
Whether a procedure is exactly coherent matters most when its incoherence ratio is near zero and its joint is implicit in its update rule, so that coherence cannot be settled by construction.
We therefore test \(H_0 : P_Y = Q_Y\), so that rejecting a heuristic procedure is evidence of incoherence with a controlled type I error.
Earlier checks of this null calibrate by bootstrap against a reference Bayesian model~\citep{falck2024incontext} or on test-set metrics~\citep{pituk2025bayesian}, whereas ours needs neither.
A batch MMD test~\citep{gretton2012kernel} would fix the number of updates in advance, and its permutation calibration fails when a round's samples share a block.
We instead adopt testing by betting~\citep{shafer2021testing}, whose anytime validity~\citep{ramdas2023game} lets the test stop once the evidence is decisive.

\subsection{Sequential kernel test of coherence}
\label{sec:sequential-test}

Our test is the kernel-MMD betting test of \citet{shekhar2023nonparametric} with its plug-in witness and the one-sided online Newton step stakes of \citet[Algorithm~1]{podkopaev2023sequential}, applied to update rounds rather than sample pairs.
With the kernel, embeddings and displacements of \Cref{sec:incoherence-ratio}, coherence implies the kernel null \(H_0^k : \mathbb{E}[\delta_r] = 0\) for any kernel, so the test is valid whatever the kernel.
The kernel determines only which departures can be detected.
A quadratic kernel, for example, tests coherence only in mean and variance.

Each round draws \(Z_r \sim P_Z\), updates the fitted procedure, and estimates \(\delta_r\) by \(\hat\delta_r\), the difference between the embeddings of \(P_Y^{Z_r}\) and \(P_Y\), computed exactly or from samples (\Cref{app:testing}).
Unlike the estimator of \(\Psi\), the test draws fresh samples of \(P_Y\) in every round.
Rounds are then independent given the fitted procedure, so \(\mathbb{E}[\hat\delta_r \mid \mathcal{F}_{r-1}] = \Delta\), where \(\mathcal{F}_{r-1}\) is everything observed before round \(r\).
A pool shared across rounds would couple the bets through its Monte Carlo error and break this property.
Before round \(r\), the test forms the running mean \(D_{r-1} := \frac{1}{r-1}\sum_{i<r} \hat\delta_i\) and the plug-in witness \(u_r := D_{r-1}/\|D_{r-1}\|\), or \(u_r := 0\) if \(\|D_{r-1}\| \le \varepsilon\) for a fixed noise floor \(\varepsilon\).
It then bets on the projection of the new displacement onto the witness,
\begin{equation}
    b_r := \frac{\langle u_r, \hat\delta_r \rangle}{2\sqrt{\kappa}},
    \qquad
    E_r := E_{r-1}\,(1 + \lambda_r b_r),
    \qquad
    E_0 := 1,
    \label{eq:wealth}
\end{equation}
where \(b_r \in [-1, 1]\) since \(\|\hat\delta_r\| \le 2\sqrt{\kappa}\), and the stake \(\lambda_r \in [0, 1/2]\) is set from earlier rounds by the online Newton step~\citep{hazan2007logarithmic} betting strategy of \citet{cutkosky2018black}.
Since \(u_r\) and \(\lambda_r\) depend only on \(\mathcal{F}_{r-1}\), \(\mathbb{E}[b_r \mid \mathcal{F}_{r-1}] = \langle u_r, \Delta \rangle / (2\sqrt{\kappa})\), which is zero under the null.
The wealth \((E_r)\) is then a non-negative martingale with \(\mathbb{E}[E_r] = 1\), and Ville's inequality~\citep{ville1939etude} gives \(\mathbb{P}_{H_0}(\sup_r E_r \ge 1/\alpha) \le \alpha\).
Rejecting the first time \(E_r \ge 1/\alpha\) therefore controls the type I error at every stopping time~\citep{shafer2011test}, and no update is spent after the evidence is decisive.

Under an alternative with \(\|\Delta\| > \varepsilon\), the witness converges to \(\Delta/\|\Delta\|\) and \(\log E_r\) grows linearly in \(r\), so the test is consistent above the noise floor, by the argument that \citet{shekhar2023nonparametric} and \citet{podkopaev2023sequential} give for their tests (\Cref{app:testing}).
Applied to rounds, their bounds on the growth rate show that it is set mainly by the absolute bias \(\|\Delta\|/(2\sqrt{\kappa})\) rather than by \(\Psi\), because bets are scaled by the worst-case bound \(2\sqrt{\kappa}\), which keeps the test valid without assumptions on the procedure but makes it conservative when updates are small (\Cref{app:severity-detectability}).
Failing to reject within a budget of updates does not certify coherence, since incoherence below the noise floor, outside the features of the kernel, or too small to accumulate evidence within the budget goes undetected.

\paragraph{Testing at multiple evaluation points.}
Coherence is local, but a procedure is used at many evaluation points, and incoherence at any of them matters.
We test the joint null that \(P_Y = Q_Y\) at every evaluation point \(l = 1, \ldots, L\).
A single update gives a displacement at every point, so testing at \(L\) points costs no more updates than testing at one.
Running the construction separately at each point gives wealths \(E_{l,r}\), which share the same updates and are dependent, so their product is not a valid test martingale.
Under dependence they are instead combined by averaging~\citep{vovk2021values}: an average of martingales with respect to a common filtration is itself a martingale, and we reject the first time \(\bar E_r := \frac{1}{L}\sum_{l} E_{l,r} \ge 1/\alpha\).
Since \(\bar E_r \ge \max_l E_{l,r}/L\), incoherence confined to a single point costs at most \(\log L\) in log-evidence relative to testing at that point alone (\Cref{app:testing}).

\section{Experiments}
\label{sec:experiments}

The first two experiments ask whether incoherence measures what it should and what it reveals about models, in a controlled sparse GP setting and across models on real tabular datasets.
The last two ask whether incoherence reaches decisions and what it costs, within mean-field Bayesian linear regression and across models on a synthetic neural-network testbed.
Throughout, \(\Psi\) is pooled over evaluation points, and tests of coherence use level \(\alpha = 0.05\).
Full experimental details are in \Cref{app:experiments}.

\begin{figure}[t]
    \centering
    \includegraphics[width=0.99\textwidth]{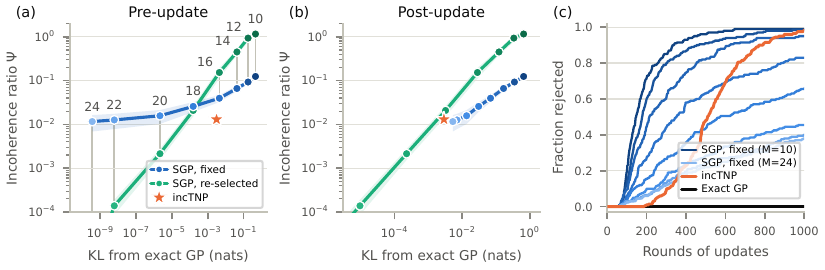}
    \caption{Sparse GPs with fixed (blue) or re-selected (green) inducing points, and an incTNP.
    \textbf{(a, b)}~\(\Psi\) against KL from the exact GP before and after the update.
    Vertical lines join procedures with the same number of inducing points, and bands show \(\pm 2\) standard errors.
    \textbf{(c)}~Fraction of tasks rejected at \(\alpha = 0.05\).
    The exact GP is never rejected.}
    \label{fig:synthetic-gp}
\end{figure}

\paragraph{Incoherence within sparse GPs.}
In a controlled sparse GP setting, the number of inducing points \(M\) controls the resolution of the approximation.
Functions are drawn from a GP prior, and each task observes \(n = 24\) points before being updated on a block of \(K = 8\) additional points, with incoherence measured at 64 evaluation points.
We draw 20 functions (environments) and form 10 tasks from each by subsampling the observed, block and evaluation points, reporting the mean over tasks.
Sparse GPs use the variational approximation of \citet{titsias2009variational}, with the kernel hyperparameters fixed at their true values.
Inducing points are chosen deterministically by greedy variance selection~\citep{burt2020convergence}, and we compare an update procedure in which the inducing points are chosen before the update and held fixed with one in which they are re-selected during the update.
The two make identical initial predictions, so they differ only in how they update.
We also include an incremental transformer neural process (incTNP)~\citep{mortimer2026incremental} trained on the same GP prior, which updates by extending its context (\Cref{app:sparse-gp}).
Within each update procedure, increasing the resolution of the approximation reduces incoherence and improves pre-update accuracy (\Cref{fig:synthetic-gp}a).
With fixed inducing points, incoherence plateaus since even when \(M=n\), the update cannot adequately represent the new data.
Re-selection avoids this, but with a fixed budget it must give up coverage of the existing data to accommodate the block, so its update departs from the current approximation, making it less coherent when \(M\) is small and more coherent when \(M\) is large.
Re-selection is nevertheless no less accurate after the update, even where it is the less coherent procedure (\Cref{fig:synthetic-gp}b), so across update procedures incoherence does not order post-update accuracy.
Rather, it registers that current and future predictions disagree, not which of them is wrong.
The incTNP is as coherent as the fixed sparse GPs with nearly all the inducing points while fitting less accurately, and after the update it is more accurate than all of them, lying close to the re-selected curve, consistent with an approximation that adapts to new data.
The test never rejects the exact GP, and within the fixed procedure rejects more coherent approximations more slowly (\Cref{fig:synthetic-gp}c).
Across families its speed to reject is not set by \(\Psi\) alone, as the incTNP is rejected far faster than fixed sparse GPs of similar \(\Psi\).

\paragraph{Incoherence across models on tabular data.}
We measure incoherence for models of several kinds on three regression and five classification datasets, over ten random splits (one for TabPFN), updating every model on random blocks of further inputs with outputs drawn from the model itself (\Cref{app:tabular}).
Exact methods have \(\Psi\) indistinguishable from zero and are never rejected.
The estimates are robust to the kernel: on the regression datasets, halving or doubling the lengthscale swaps only models whose \(\Psi\) is within \(\sim 15\%\) of each other.
\(\Psi\) responds to how a model updates, even when its current predictions are unchanged.
Refitting a GP's hyperparameters at each update leaves its current predictions identical to the exact GP's but introduces a small incoherence, as \citet{mlodozeniec2024implicitly} find for exchangeability.
Refitting TabPFN's preprocessing on the updated context changes its \(\Psi\) by up to a factor of six at similar log loss, although the two variants also differ before the update.
An approximately Bayesian provenance is also no guarantee of coherence.
For example, the update in BART reruns a sampler that mixes poorly within a run, and its median \(\Psi\) of 0.71 is comparable to the regression medians of nearest neighbours (0.47) and CatBoost (0.90).
Plug-in logistic regression, by contrast, is about as coherent as the Laplace approximations, with a median \(\Psi\) below 0.01.
The ordering of models is largely shared across datasets, with rank correlations of at least 0.88 between them, but the level of \(\Psi\) is not (kernel ridge ranges from 0.008 to 0.12), so \(\Psi\) ranks models only within a problem.
Finally, accuracy does not imply coherence.
TabPFN, which departs from the martingale condition~\citep{ng2026tabmgp}, has the lowest log loss on seven of the eight datasets in one variant or the other, yet on every classification dataset its \(\Psi\) is an order of magnitude or more above that of the full Laplace approximation, which is nearly coherent but generally less accurate (\Cref{tab:tabular-summary}).

\begin{figure}[t]
    \centering
    \includegraphics[width=0.99\textwidth]{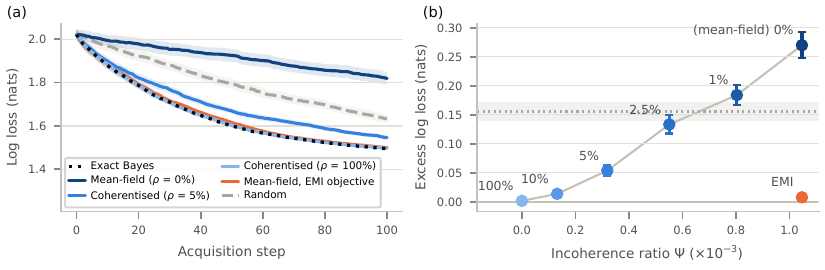}
    \caption{Acquisition with mean-field Bayesian linear regression.
    Every curve is the test log loss of the mean-field model on the data it acquired.
    \textbf{(a)}~Test log loss by acquisition step.
    \textbf{(b)}~Excess log loss over acquisition with the exact predictive, averaged over steps, against \(\Psi\).
    Points are labelled by the share \(\rho\) of the pool over which the acquisition predictive is made coherent, with \(\rho = 0\%\) the mean-field model with \(\EPIG_H\).
    EMI changes only the objective, so it is plotted at the mean-field model's \(\Psi\).
    Bands and error bars show \(\pm 2\) standard errors.}
    \label{fig:blr}
\end{figure}

\paragraph{Incoherence and acquisition within mean-field BLR.}
Incoherence means that a model's current and future predictions disagree, and we now ask whether that disagreement reaches decisions.
To isolate its effect, we hold the model and its predictions fixed and vary only the coherence of what the acquisition function consumes.
We use Bayesian linear regression on random Fourier features and compare exact Bayes with mean-field variational inference, which we compute analytically so that any incoherence comes from the approximation rather than from optimisation error.
Starting from 25 points, each model acquires 100 further points one at a time greedily from a pool, choosing each with either the entropy form of EPIG or EMI, and we average over 100 environments.
With \(\EPIG_H\), the mean-field model acquires worse data than random acquisition (\Cref{fig:blr}a).
With EMI, the same model nearly matches exact Bayes, with an excess of 0.008 nats, so the difference lies entirely in the objective.
EMI is tractable here, but in general it requires the entropy of a mixture, so we also consider making the predictive itself coherent, building a ladder in which the mean-field predictive is made coherent with its own refit over a growing random share \(\rho\) of the pool (\Cref{app:blr}).
Excess loss falls monotonically with \(\Psi\) along the ladder, and coherentising over a small random subset already recovers most of the gap to exact Bayes (\Cref{fig:blr}b).
This suggests that coherence need only hold over a representative selection of the pool rather than over every point the model might acquire.

\begin{figure}[t]
    \centering
    \includegraphics[width=0.99\textwidth]{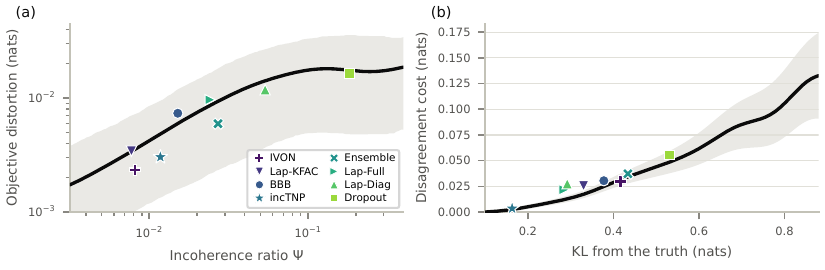}
    \caption{Acquisition on the neural testbed for eight approximate models.
    Exact Bayes is omitted, as its objectives never disagree.
    \textbf{(a)}~Distortion of the acquisition objective against \(\Psi\).
    \textbf{(b)}~Cost of a disagreement, the excess loss of \(\EPIG_H\)'s choice over EMI's, against the model's KL from the truth before the update.
    Curves are local linear fits, markers are model means, and bands show the 10th--90th percentiles in (a) and \(\pm 2\) standard errors in (b).
    BBB is Bayes by backprop.}
    \label{fig:neural-testbed}
\end{figure}

\paragraph{Incoherence and acquisition across models.}
We now ask how incoherence affects acquisition across different approximate models in a more realistic setting.
We use a synthetic testbed modelled on the Neural Testbed of \citet{osband2022neural}, with classification tasks generated by random networks, in which choosing one of several candidate batches to acquire is small enough that every labelling can be enumerated (\Cref{app:neural-testbed}).
We consider eight approximate models (Laplace approximations, variational methods, deep ensembles, MC dropout and an incTNP) over 1000 environments.
We measure the cost of a disagreement between objectives as the excess loss of \(\EPIG_H\)'s choice over EMI's, and the distortion of the objective as the spread of \({\EPIG_H} - {\EMI}\) across candidates.
Exact Bayes over a large discrete bank of networks, included as a coherent baseline, never separates the objectives, whereas for every approximate model the entropy form of EPIG disagrees with EMI on 26--80\% of decisions and has a higher loss on average.
The KL form of EPIG disagrees with EMI on only 1--20\% of decisions, as the identity of \Cref{sec:objective-splitting} predicts, making it a potentially tractable substitute for EMI when a model is incoherent.
\(\Psi\) tracks how far incoherence distorts the objective (\Cref{fig:neural-testbed}a), and within an environment the ranking of models by \(\Psi\) has a rank correlation of 0.73 with their distortion.
This makes \(\Psi\) a measure of how susceptible a model's decisions are to incoherence.
It does not tell us whether a choice will flip, which also depends on the margin between the best candidates, or what a flip will cost, which is uncorrelated with \(\Psi\) and instead tracks the model's accuracy (\Cref{fig:neural-testbed}b, \Cref{app:neural-testbed}).
Bayes by backprop and the incTNP, for example, are about equally incoherent, yet the entropy form of EPIG disagrees with EMI on 80\% of Bayes by backprop's decisions against 26\% of the incTNP's, and a disagreement costs 0.030 against 0.0035 nats.

\section{Discussion}
\label{sec:discussion}

Many sequential decision procedures act as though a model's current and future beliefs are consistent.
Measuring incoherence quantifies how far approximate models violate this, and flags when a procedure's decisions are exposed to it.
Our experiments show this is more than a theoretical concern, as acquisition objectives that should coincide frequently disagreed and chose different data.
Incoherence indicates when this disagreement can occur, but not what it costs, which depends on accuracy.
Susceptibility can also depend on the objective.
On the neural testbed, the KL form of EPIG stayed much closer to EMI than the entropy form, suggesting it as a cheap way to mitigate some of the effect of incoherence in BED.
Whilst our experiments are restricted to BED, our framework applies to any single-update sequential decision problem.
Understanding which properties of a problem determine how much incoherence matters would show where these diagnostics are most needed.
Extending the framework to procedures that plan through a sequence of actions and updates would cover longer-horizon problems, where the effects of incoherence may compound.
Beyond the choice of decision problem, the main limitations of measuring incoherence are the computational cost of many model updates and the scope of what it can tell us.
The best models are both accurate and coherent, but coherence alone guarantees little, since exact conditioning in a misspecified model is perfectly coherent and can still be badly wrong.
Measuring incoherence can reveal a potential problem, and help in explaining and addressing it, but it cannot substitute for held-out evaluation when choosing a model.
Once an accurate model is chosen, making it exactly coherent is currently intractable, and approximate coherentisation over a representative subset of the pool has so far been shown to help only in a simple setting.
It nonetheless points to a promising direction.
Models that excel at static prediction are rarely designed for sequential decisions, and approaches that account for incoherence could help them perform well there too.

\begin{ack}
JR was supported by the EPSRC Centre for Doctoral Training in Statistics and Machine Learning (StatML), grant EP/Y034813/1.
VS was supported by the EPSRC Centre for Doctoral Training in Modern Statistics and Statistical Machine Learning (StatML), grant EP/S023151/1.
The authors declare no competing interests.
\end{ack}

\bibliographystyle{unsrtnat}
\bibliography{references}

\clearpage
\raggedbottom
\appendix
\crefalias{section}{appendix}
\crefalias{subsection}{appendix}

\section{Theoretical foundations}
\label{app:foundations}

\subsection{Diachronic Dutch book}
\label{app:dutch-book}

We follow the diachronic Dutch book of \citet{teller1973conditionalization} and \citet{lewis1999conditionalize}, in the notation of \Cref{sec:coherence}.
Fix the history \(h\), the target \(Y\) and the update design \(d\) with block \(Z\).
The procedure acts as an agent who will buy or sell any contract at its fair price, the contract's expected payoff under the agent's predictive at the history where it is offered.
For a payoff \(f : \mathcal{Y} \to [0,1]\), paid once \(Y\) is observed, the fair prices before and after the update are
\begin{equation*}
    \mathcal{V}_0(f) = \mathbb{E}_{P_Y}[f(Y)],
    \qquad
    \mathcal{V}_z(f) = \mathbb{E}_{P_Y^z}[f(Y)].
\end{equation*}
Before the update, a contract paying the future price \(\mathcal{V}_Z(f)\) once \(Z\) is revealed has fair price
\begin{equation*}
    \overline{\mathcal{V}}(f) = \mathbb{E}_{P_Z}\!\left[\mathcal{V}_Z(f)\right] = \mathbb{E}_{Q_Y}[f(Y)],
\end{equation*}
where the second equality follows from~\eqref{eq:expected-future-predictive}.

\paragraph{Construction.}
Suppose \(\Delta(f) := \mathcal{V}_0(f) - \overline{\mathcal{V}}(f) > 0\).
If instead \(\Delta(f) < 0\), replace \(f\) by \(1 - f\), which reverses the sign.
At a stake \(\alpha > 0\), a bookmaker executes three trades, each at the agent's price.
\begin{enumerate}[label=(\Alph*)]
    \item At \(t=0\), sell the agent a contract paying \(\alpha f(Y)\) at \(t=2\), for \(\alpha\mathcal{V}_0(f)\).
    \item At \(t=0\), buy from the agent a contract paying \(\alpha\mathcal{V}_Z(f)\) at \(t=1\), for \(\alpha\overline{\mathcal{V}}(f)\).
    \item At \(t=1\), after \(Z=z\) is observed, buy from the agent a contract paying \(\alpha f(Y)\) at \(t=2\), for \(\alpha\mathcal{V}_z(f)\).
\end{enumerate}
Starting from wealth 1, the agent holds \(W_0 = 1 - \alpha\Delta(f)\) after the trades at \(t=0\).
At \(t=1\) it pays \(\alpha\mathcal{V}_z(f)\) to settle B and receives the same amount for C, so \(W_1 = W_0\).
At \(t=2\) the payoffs of A and C cancel, so \(W_2 = 1 - \alpha\Delta(f)\) whatever the values of \(Z\) and \(Y\), a sure loss of \(\alpha\Delta(f)\).

\paragraph{Bounded liability.}
At every time, the agent's wealth must cover the sum of the largest possible payments on the contracts it has sold, without offsetting the contracts it holds.
In the construction the agent has sold only B after \(t=0\) and only C after \(t=1\), each paying at most \(\alpha\), so the constraint holds whenever \(1 - \alpha\Delta(f) \geq \alpha\), that is, \(\alpha \leq 1/(1 + \Delta(f))\).
At \(\alpha = 1/(1 + \Delta(f))\) the sure loss is
\begin{equation}
    1 - W_2 = \frac{\Delta(f)}{1 + \Delta(f)}.
    \label{eq:app-dutch-loss}
\end{equation}

\paragraph{Maximal sure loss.}
The loss~\eqref{eq:app-dutch-loss} increases with \(\Delta(f)\), whose supremum over \(f\) is \(D_{\TV} := \TV(P_Y, Q_Y)\).
The construction therefore extracts a fraction \(D_{\TV}/(1 + D_{\TV})\) of initial wealth, and no book does better.
Consider any book of trades at the agent's prices, in contracts on \(Y\) at \(t=0\) and \(t=1\) and on \(Z\) at \(t=0\), and let \(W_1\) be the agent's wealth after the trades at \(t=1\).
Write \(f_0\) and \(f_z\) for the agent's net payoffs from the contracts on \(Y\) traded at \(t=0\) and at \(t=1\) after \(Z=z\), so that \(W_2 = W_1 + f_0(Y) + f_z(Y)\).
Split them into contracts held and sold, \(f_0 = u_0 - s_0\) and \(f_z = u_z - s_z\) with \(u_0, u_z, s_0, s_z \geq 0\).
Every trade is fair, so averaging \(W_2\) over \(Y \sim P_Y^z\) and then \(Z \sim P_Z\) gives \(1 - \mathbb{E}_{P_Y}[f_0] + \mathbb{E}_{Q_Y}[f_0]\).
A sure loss \(L\), with \(W_2 \leq 1 - L\) whatever the values of \(Z\) and \(Y\), therefore satisfies
\begin{equation*}
    L \leq \mathbb{E}_{P_Y}[f_0] - \mathbb{E}_{Q_Y}[f_0] \leq D_{\TV}\,(\sup f_0 - \inf f_0).
\end{equation*}
Bounded liability after the trades at \(t=1\) requires \(W_1 \geq \sup s_0 + \sup s_z\).
Since \(W_2 \leq 1 - L\) and \(-f_z \leq s_z\), this gives \(f_0 = W_2 - W_1 - f_z \leq 1 - L - W_1 + \sup s_z \leq 1 - L - \sup s_0\), while \(f_0 \geq -\sup s_0\).
Hence \(\sup f_0 - \inf f_0 \leq 1 - L\), and \(L \leq D_{\TV}(1 - L)\) gives \(L \leq D_{\TV}/(1 + D_{\TV})\).
The maximal sure loss vanishes if and only if \(P_Y = Q_Y\), that is, exactly when the procedure is coherent for \(Y\), \(h\) and \(Z\).

\subsection{Forms of EPIG}
\label{app:epig-forms}

This section derives the identities~\eqref{eq:epig-kl-identity} and~\eqref{eq:epig-h-identity} relating the three forms of EPIG, and shows why the KL form stays closer to EMI than the entropy form.
Fix the history \(h\) and the candidate design \(a\), and write \(p_Y\) and \(q_Y^a\) for the densities of \(P_Y\) and \(Q_Y^a\).
We assume that \(P_Y^{a,z} \ll P_Y\) for \(P_Z^a\)-almost every \(z\), and that all entropies and cross-entropies below are finite.

\paragraph{Relating the three forms.}
Since \(Q_Y^a\) averages the post-update predictives, \(Q_Y^a = \mathbb{E}_{Z \sim P_Z^a}[P_Y^{a,Z}]\), Fubini's theorem gives \(\mathbb{E}_{Z \sim P_Z^a}\,\mathbb{E}_{P_Y^{a,Z}}[\log p_Y] = \mathbb{E}_{Q_Y^a}[\log p_Y]\).
Expanding the KL divergence inside \(\EPIG_{\KL}\) therefore gives
\begin{equation*}
    \EPIG_{\KL}(a) = -\mathbb{E}_{Z \sim P_Z^a}\big[H[P_Y^{a,Z}]\big] - \mathbb{E}_{Q_Y^a}[\log p_Y].
\end{equation*}
Substituting \(-\mathbb{E}_{Q_Y^a}[\log p_Y] = H[Q_Y^a] + \KL(Q_Y^a \,\|\, P_Y)\) yields~\eqref{eq:epig-kl-identity}.
Identity~\eqref{eq:epig-h-identity} is immediate, since \(\EPIG_H\) and EMI share the term \(\mathbb{E}_{Z \sim P_Z^a}\big[H[P_Y^{a,Z}]\big]\).
Under coherence, \(Q_Y^a = P_Y\), the extra terms in both identities vanish, and all three forms equal EMI.
Substituting \(Q_Y^a\) for \(P_Y\) in \(\EPIG_H\) or \(\EPIG_{\KL}\) gives EMI, which does not involve \(P_Y\), so all three forms coherentise to EMI.

\paragraph{Size of the departures.}
Write \(q_Y^a = p_Y(1 + \eta g_a)\), where \(g_a\) is bounded with \(\mathbb{E}_{P_Y}[g_a] = 0\), so that \(\eta\) sets the size of the incoherence.
Expanding \((1+x)\log(1+x) = x + x^2/2 + O(x^3)\) gives
\begin{equation*}
    \KL(Q_Y^a \,\|\, P_Y) = \tfrac{\eta^2}{2}\,\mathbb{E}_{P_Y}[g_a^2] + O(\eta^3) = \tfrac{1}{2}\chi^2(Q_Y^a \,\|\, P_Y) + O(\eta^3),
\end{equation*}
so the KL form departs from EMI only at second order.
The entropy form departs from EMI by \(H[P_Y] - H[Q_Y^a] = \KL(Q_Y^a \,\|\, P_Y) + \mathbb{E}_{Q_Y^a}[\log p_Y] - \mathbb{E}_{P_Y}[\log p_Y]\), whose last two terms equal \(\eta\,\mathbb{E}_{P_Y}[g_a \log p_Y]\), so it is first order and can take either sign.

\paragraph{Consequence for rankings.}
Only differences between candidates affect which design is chosen.
For a finite candidate set, let \(a^+\) maximise EMI, and let \(m := \min_{a \neq a^+}\{\EMI(a^+) - \EMI(a)\}\) be its margin over the next best candidate.
Since KL divergence is non-negative, identity~\eqref{eq:epig-kl-identity} gives
\begin{equation*}
    \EPIG_{\KL}(a^+) - \EPIG_{\KL}(a) \geq m - \KL(Q_Y^a \,\|\, P_Y)
    \qquad \text{for every } a \neq a^+,
\end{equation*}
so the KL form selects the same design as EMI whenever \(m > \max_{a \neq a^+} \KL(Q_Y^a \,\|\, P_Y)\), a second-order quantity.
By identity~\eqref{eq:epig-h-identity}, the same argument guarantees agreement for the entropy form only when \(m > \max_{a \neq a^+}\{H[Q_Y^{a^+}] - H[Q_Y^a]\}\), which is first order.

\section{Measuring and testing incoherence}
\label{app:ratio}

We use the notation of \Cref{sec:incoherence-ratio}, with a kernel \(k\) on \(\mathcal{Y}\) whose feature map \(\phi\) has \(\|\phi(y)\|^2 \le \kappa\).
Round \(r\) draws \(Z_r \sim P_Z\) and returns the post-update predictive \(P_Y^{Z_r}\), whose mean embedding~\citep{muandet2017kernel} we write \(\mu_r\), and
\begin{align*}
    \delta_r &:= \mu_r - \mu_P, \\
    \Delta &:= \mu_Q - \mu_P = \mathbb{E}[\delta_r], \\
    V &:= \mathbb{E}\big[\|\delta_r - \Delta\|^2\big] = \mathbb{E}\big[\|\mu_r - \mu_Q\|^2\big].
\end{align*}
Each round updates the same fitted procedure, drawing the block and any randomness internal to the update afresh.
Since \(Q_Y\) averages over both, \(\mu_1, \ldots, \mu_R\) are independent and identically distributed with mean \(\mu_Q\).

\subsection{Estimation}
\label{app:ratio-estimation}

Every estimator below is a U-statistic~\citep{hoeffding1948class}, an average over pairs of distinct rounds and, when embeddings are estimated from samples, of distinct samples.
As for the unbiased MMD estimator of \citet{gretton2012kernel}, excluding self-pairs is what makes each one unbiased.

\paragraph{Exact embeddings.}
The numerator and denominator are order-two U-statistics over rounds,
\begin{align*}
    \hat N &:= \frac{1}{R(R-1)}\sum_{r \neq r'} \langle \delta_r, \delta_{r'} \rangle, \\
    \hat V &:= \frac{1}{2R(R-1)}\sum_{r \neq r'} \|\mu_r - \mu_{r'}\|^2,
\end{align*}
and \(\hat\Psi := \hat N / \hat V\).
For \(r \neq r'\), independence of the rounds gives \(\mathbb{E}[\langle \delta_r, \delta_{r'} \rangle] = \|\Delta\|^2\) and \(\mathbb{E}\big[\tfrac{1}{2}\|\mu_r - \mu_{r'}\|^2\big] = V\), so both are unbiased.
With \(\hat\Delta := \frac{1}{R}\sum_r \delta_r\), they equal
\begin{align*}
    \hat V &= \frac{1}{R-1}\sum_{r=1}^R \|\delta_r - \hat\Delta\|^2, \\
    \hat N &= \|\hat\Delta\|^2 - \hat V/R,
\end{align*}
the sample variance of the displacements and a plug-in corrected for \(\mathbb{E}\big[\|\hat\Delta\|^2\big] = \|\Delta\|^2 + V/R\), since the covariance of \(\hat\Delta\) has trace \(V/R\).

\paragraph{Sampled embeddings.}
Round \(r\) provides \(S\) independent samples \(y_{r,1}, \ldots, y_{r,S}\) from \(P_Y^{Z_r}\), independently across rounds, and every round shares one reference sample of \(T\) independent draws \(y'_1, \ldots, y'_T\) from \(P_Y\).
Define
\begin{align*}
    K_{rr'} &:= \frac{1}{S^2}\sum_{s,s'} k(y_{r,s}, y_{r',s'}) \quad (r \neq r'), \\
    \tilde K_{r} &:= \frac{1}{S(S-1)}\sum_{s \neq s'} k(y_{r,s}, y_{r,s'}), \\
    K_{rP} &:= \frac{1}{ST}\sum_{s,t} k(y_{r,s}, y'_t), \\
    K_{P} &:= \frac{1}{T(T-1)}\sum_{t \neq t'} k(y'_t, y'_{t'}).
\end{align*}
None of these pairs a sample with itself, so, given \(P_Y^{Z_1}, \ldots, P_Y^{Z_R}\), they are unbiased for \(\langle \mu_r, \mu_{r'} \rangle\), \(\|\mu_r\|^2\), \(\langle \mu_r, \mu_P \rangle\) and \(\|\mu_P\|^2\) respectively.
Replacing each inner product in the exact-embedding estimators by the corresponding kernel average gives
\begin{align*}
    \hat N &:= \frac{1}{R(R-1)}\sum_{r \neq r'} K_{rr'} - \frac{2}{R}\sum_r K_{rP} + K_P, \\
    \hat V &:= \frac{1}{R}\sum_r \tilde K_r - \frac{1}{R(R-1)}\sum_{r \neq r'} K_{rr'}, \\
    \hat\Psi &:= \hat N / \hat V.
\end{align*}
Given the predictives, \(\hat N\) and \(\hat V\) are therefore unbiased for their exact-embedding counterparts, and hence unbiased.
The sampled \(\hat V\) can be negative, and its Monte Carlo error includes a term of order \(\kappa/(S\sqrt{R})\), which must be small relative to \(V\) for \(\hat\Psi\) to be reliable.

\paragraph{Reusing the reference sample.}
The reference sample enters only \(\hat N\), through \(K_{rP}\) and \(K_P\).
Sharing it across rounds correlates the terms of \(\hat N\) but introduces no bias.
Its Monte Carlo error, however, does not shrink with \(R\): as \(R \to \infty\), \(\hat N\) still fluctuates about \(\|\Delta\|^2\) by order \(\|\Delta\|\,T^{-1/2} + T^{-1}\), against \(\|\Delta\|\,(RS)^{-1/2}\) from the round samples.
The reference sample should therefore be large relative to \(RS\), which is cheap because sampling \(P_Y\) needs no update, and comparing \(\hat N\) across its two halves checks this.

\paragraph{Cost.}
With an explicit feature map of dimension \(D\), such as the Dirac kernel or random Fourier features~\citep{rahimi2007random}, each round reduces to its embedding in \(\mathbb{R}^D\), exact or averaged over \(S\) samples.
The point estimates then cost \(O(RD)\), or \(O((RS + T)D)\) with samples, in memory independent of \(R\), although the standard error below needs all \(R\) embeddings.
With only a kernel, they compare every pair of rounds, at cost \(O(R^2)\) with exact embeddings and \(O((RS + T)^2)\) with samples.

\paragraph{Pooling over evaluation points.}
With estimators \(\hat N_l\) and \(\hat V_l\) at evaluation points \(l = 1, \ldots, L\), we report
\begin{equation*}
    \hat\Psi := \frac{\sum_l \hat N_l}{\sum_l \hat V_l},
\end{equation*}
which estimates \(\Psi\) under the direct-sum kernel over points and so keeps its scale invariance.
This pooled \(\Psi\) is the \(V_l\)-weighted average of the per-point ratios \(\Psi_l := \|\Delta_l\|^2 / V_l\), whose estimates \(\hat N_l / \hat V_l\) localise incoherence.

\paragraph{Standard errors.}
With \(\hat\Psi_{(-r)}\) the estimate without round \(r\), and \(\bar\Psi\) the mean of these \(R\) values, the leave-one-round-out standard error
\begin{equation*}
    \mathrm{SE} := \bigg(\frac{R-1}{R}\sum_{r=1}^R \big(\hat\Psi_{(-r)} - \bar\Psi\big)^2\bigg)^{1/2}
\end{equation*}
captures the variability of numerator and denominator jointly.
With sampled embeddings it holds the reference sample fixed, so it omits that sample's Monte Carlo error, which is negligible when \(T \gg RS\).
Since \(\hat N\) and \(\hat V\) are unbiased, the ratio has an \(O(R^{-1})\) bias by the delta method~\citep{vandervaart1998asymptotic}.

\paragraph{Behaviour near coherence.}
When \(\Delta = 0\), \(\mathbb{E}[\langle \delta_r, \delta_{r'} \rangle \mid \delta_r] = \langle \delta_r, \Delta \rangle\) vanishes, so \(\hat N\) is a degenerate U-statistic~\citep{serfling1980approximation}.
It is then of order \(R^{-1}\) rather than \(R^{-1/2}\): \(R\hat N\) converges in distribution to a centred weighted sum of \(\chi^2_1\) variables, and \(\hat N\) is negative with appreciable probability.
We report \(\hat N\) and \(\hat\Psi\) unclipped, so that averaging over tasks does not bias them upwards, and read negative values as indistinguishable from zero.
The standard error then also falls as \(R^{-1}\) (\Cref{tab:sgp-rounds}), but it is not consistent: the leave-one-round-out variance of the degenerate \(\hat N\) has expectation \(2(R-1)/(R-2)\) times its true variance, so the standard error overstates the spread of \(\hat\Psi\) by a factor of about \(\sqrt{2}\).
It remains a useful, conservative scale, but, as \Cref{sec:testing-coherence} notes, \(\hat\Psi \pm 2\,\mathrm{SE}\) is not a confidence interval for a procedure that may be exactly coherent.

\subsection{Properties}
\label{app:ratio-properties}

\paragraph{Scale invariance and the systematic share.}
Rescaling every displacement, \(\delta_r \mapsto c\,\delta_r\) with \(c > 0\), multiplies \(\|\Delta\|^2\) and \(V\) by \(c^2\) and so leaves \(\Psi\) unchanged.
Since \(\mathbb{E}\big[\|\delta_r\|^2\big] = \|\Delta\|^2 + V\),
\begin{equation*}
    \frac{\Psi}{1+\Psi} = \frac{\|\Delta\|^2}{\|\Delta\|^2 + V} = \frac{\|\Delta\|^2}{\mathbb{E}\big[\|\delta_r\|^2\big]} \in [0, 1]
\end{equation*}
is the systematic share of the expected squared displacement.

\paragraph{The limit \(V \to 0\).}
If \(V = 0\), then \(\mu_r = \mu_Q\) almost surely, so for a characteristic kernel~\citep{sriperumbudur2010hilbert} \(P_Y^{Z_r} = Q_Y\) almost surely.
If \(\Delta \neq 0\), the update is a shift that ignores the block, which is wholly systematic, so \(\Psi = \infty\).
If \(\Delta = 0\), the update leaves the predictive unchanged, which is trivially coherent, and we set \(\Psi := 0\), so that, for a characteristic kernel, \(\Psi = 0\) if and only if the procedure is coherent.
For small positive \(V\), \(\Psi\) still reports whether what little movement there is tracks the block or drifts systematically, and \(\hat N\) and \(\hat V\) give the absolute sizes of the bias and the spread.

\paragraph{Randomised updates.}
Let \(\xi_r\) be the randomness internal to the update in round \(r\), so that \(\delta_r = \delta(Z_r, \xi_r)\), and let \(\bar\delta(z) := \mathbb{E}[\delta(z, \xi)]\) be the mean displacement for block \(z\).
The law of total variance gives \(V = V_Z + V_\xi\), where
\begin{align*}
    V_Z &:= \mathbb{E}\big[\|\bar\delta(Z_r) - \Delta\|^2\big] \quad \text{(spread driven by the block)}, \\
    V_\xi &:= \mathbb{E}\big[\|\delta_r - \bar\delta(Z_r)\|^2\big] \quad \text{(update noise)}.
\end{align*}
Randomising the update without changing its average for each block leaves \(Q_Y\), and hence \(\Delta\) and the Dutch book, unchanged, while increasing \(V\) and lowering \(\Psi\).
By convexity of \(-\log\), it also cannot improve the expected log score, so update noise is not a free way to appear coherent.
To exclude update noise, use the block-driven ratio
\begin{equation*}
    \Psi_Z := \frac{\|\Delta\|^2}{V_Z} \ge \Psi,
\end{equation*}
which equals \(\Psi\) for deterministic updates.
It costs a second update per block: for two updates \(\delta_r^{(1)}\) and \(\delta_r^{(2)}\) on the same block with independent internal randomness, \(\mathbb{E}\big[\langle \delta_r^{(1)}, \delta_r^{(2)} \rangle\big] = \|\Delta\|^2 + V_Z\), so
\begin{equation*}
    \hat V_Z := \frac{1}{R}\sum_{r=1}^R \langle \delta_r^{(1)}, \delta_r^{(2)} \rangle - \hat N
\end{equation*}
is unbiased for \(V_Z\), with \(\hat N\) computed from the \(\delta_r^{(1)}\).

\paragraph{Embedding discrepancy and total variation.}
Every \(f\) in the unit ball of the RKHS satisfies \(|f(y)| = |\langle f, \phi(y) \rangle| \le \sqrt{\kappa}\), and \(|\mathbb{E}_{P_Y}[f] - \mathbb{E}_{Q_Y}[f]| \le 2\|f\|_\infty D_{\TV}\), so
\begin{equation*}
    \|\Delta\| = \MMD(P_Y, Q_Y) = \sup_{\|f\| \le 1} \big|\mathbb{E}_{P_Y}[f] - \mathbb{E}_{Q_Y}[f]\big| \le 2\sqrt{\kappa}\, D_{\TV}.
\end{equation*}
The sure loss \(D_{\TV}/(1 + D_{\TV})\) of \Cref{app:dutch-book} increases with \(D_{\TV}\), so the embedding discrepancy bounds it from below, by \(\|\Delta\|/(2\sqrt{\kappa} + \|\Delta\|)\), but gives no upper bound in general.
For a finite \(\mathcal{Y}\) with the Dirac kernel, \(\phi(y) = e_y\), so \(\|\Delta\| = \|q - p\|_2\) and \(D_{\TV} = \tfrac{1}{2}\|q - p\|_1\), where \(p\) and \(q\) are the probability vectors of \(P_Y\) and \(Q_Y\).
Since \(q - p\) sums to zero, its positive and negative parts each have \(\ell_1\)-norm \(D_{\TV}\), so \(\|q - p\|_2^2 \le 2 D_{\TV}^2\).
With \(\|q - p\|_2 \ge \|q - p\|_1/\sqrt{|\mathcal{Y}|}\), this gives
\begin{equation*}
    \frac{2 D_{\TV}}{\sqrt{|\mathcal{Y}|}} \le \|\Delta\| \le \sqrt{2}\, D_{\TV},
\end{equation*}
so in the classification experiments the two agree up to a factor of \(\sqrt{|\mathcal{Y}|/2}\).

\subsection{The sequential test}
\label{app:testing}

The difference between the features of a pair of samples in the test of \citet{shekhar2023nonparametric} becomes the displacement estimate \(\hat\delta_r\), and their witness of norm \(1/2\) under a kernel bounded by one becomes \(u_r/(2\sqrt{\kappa})\).

\paragraph{Displacement estimates.}
With exact embeddings, \(\hat\delta_r := \delta_r\).
Otherwise, with an explicit feature map, round \(r\) draws \(S\) samples \(y_{r,1}, \ldots, y_{r,S}\) from \(P_Y^{Z_r}\) and a fresh set of \(T\) samples \(y'_{r,1}, \ldots, y'_{r,T}\) from \(P_Y\), and sets
\begin{equation*}
    \hat\delta_r := \frac{1}{S}\sum_{s=1}^S \phi(y_{r,s}) - \frac{1}{T}\sum_{t=1}^T \phi(y'_{r,t}).
\end{equation*}
Each embedding or average of features has norm at most \(\sqrt{\kappa}\), so \(\|\hat\delta_r\| \le 2\sqrt{\kappa}\) and every bet lies in \([-1, 1]\).
All our experiments use exact embeddings with \(\kappa = 1\): the Dirac kernel has \(\phi(y) = e_y\), and random Fourier features are normalised so that \(\|\phi(y)\|^2 = 1\).

\paragraph{Why the test needs fresh samples.}
Given the fitted procedure, each round draws the block, the update's internal randomness and both sets of samples afresh, so \(\hat\delta_r\) is independent of \(\mathcal{F}_{r-1}\) and \(\mathbb{E}[\hat\delta_r \mid \mathcal{F}_{r-1}] = \Delta\).
If every round instead reused one reference sample, its estimate \(\hat\mu_P\) would be fixed by the past, so \(\mathbb{E}[\hat\delta_r \mid \mathcal{F}_{r-1}] = \mu_Q - \hat\mu_P\), which under the null is the Monte Carlo error of \(\hat\mu_P\) rather than zero.
The witness would learn this drift, and the test would eventually reject a coherent procedure.
The batch estimator of \Cref{app:ratio-estimation} needs only unconditional unbiasedness, so it can share one reference sample, although that sample's error then does not average out.

\paragraph{Witness and noise floor.}
The witness \(u_r\) of \Cref{sec:sequential-test} is the plug-in estimate that \citet{shekhar2023nonparametric} describe for their test and \citet{podkopaev2023sequential} use for independence testing, plus our noise floor.
We use \(\varepsilon = 10^{-8}\), which guards against normalising a numerically zero vector, and \(D_0 := 0\), so the first round never bets.
Any rule for \(u_r\) that depends only on \(\mathcal{F}_{r-1}\) and has \(\|u_r\| \le 1\) preserves validity, so the floor affects power but not the type I error.

\paragraph{Stakes.}
The stake is chosen by online Newton steps on the log-wealth~\citep{hazan2007logarithmic}, proposed for betting by \citet{cutkosky2018black}, in the one-sided form of \citet[Algorithm~1]{podkopaev2023sequential}.
Each evaluation point keeps its own stake and curvature, initialised at \(\lambda_1 := 0\) and \(A_0 := 1\).
After round \(r\), with \(b_r\) observed and the wealth already updated using \(\lambda_r\),
\begin{align*}
    z_r &:= \frac{b_r}{1 + \lambda_r b_r}, \\
    A_r &:= A_{r-1} + z_r^2, \\
    \lambda_{r+1} &:= \clip_{[0, 1/2]}\Big(\lambda_r + c\,\frac{z_r}{A_r}\Big),
    \qquad c := \frac{2}{2 - \log 3} \approx 2.22,
\end{align*}
where \(z_r\) is the derivative of \(\log(1 + \lambda b_r)\) in \(\lambda\) at \(\lambda_r\).
In one dimension, clipping to \([0, 1/2]\) coincides with projection in the curvature norm.
The stake \(\lambda_{r+1}\) depends only on rounds up to \(r\), so it is predictable, and the cap at \(1/2\) keeps \(1 + \lambda_r b_r \ge 1/2\), as the logarithmic regret guarantee of online Newton steps against the best constant stake requires.

\paragraph{Validity.}
Fix an evaluation point.
Under the null, \(\mathbb{E}[\hat\delta_r \mid \mathcal{F}_{r-1}] = 0\), and \(u_r\) and \(\lambda_r\) are \(\mathcal{F}_{r-1}\)-measurable, so the bounded bets \(b_r\) have conditional mean zero~\citep{waudby2024estimating} and
\begin{equation*}
    \mathbb{E}[E_r \mid \mathcal{F}_{r-1}] = E_{r-1}\big(1 + \lambda_r\, \mathbb{E}[b_r \mid \mathcal{F}_{r-1}]\big) = E_{r-1}.
\end{equation*}
Since \(1 + \lambda_r b_r \ge 1/2\), \((E_r)\) is a positive martingale with \(E_0 = 1\).
Ville's inequality, \(\mathbb{P}_{H_0}(\sup_r E_r \ge 1/\alpha) \le \alpha\), then bounds the type I error by \(\alpha\) at any stopping time.

\paragraph{Multiple evaluation points.}
Let \(\mathcal{F}_r\) contain everything observed at every evaluation point up to round \(r\).
By the argument above, each per-point wealth \((E_{l,r})\) is a non-negative martingale with respect to this common filtration, whatever the dependence between points, so their average \(\bar E_r\)~\citep{vovk2021values} is one too, with \(\bar E_0 = 1\), so Ville's inequality applies to it.
Equivalently, \(\bar E_R\) is a product of predictable convex combinations of the per-point factors,
\begin{equation*}
    \bar E_R = \prod_{r=1}^R \sum_{l=1}^L w_{l,r}\,(1 + \lambda_{l,r} b_{l,r}),
    \qquad
    w_{l,r} := \frac{E_{l,r-1}}{\sum_{l'=1}^L E_{l',r-1}},
\end{equation*}
since each factor equals \(\sum_l E_{l,r} / \sum_l E_{l,r-1}\) and the product telescopes.
Since every \(E_{l,r}\) is positive, \(\bar E_r \ge \max_l E_{l,r}/L\), so \(\log \bar E_r \ge \max_l \log E_{l,r} - \log L\).
We compute \(\log \bar E_r\) with a log-sum-exp over points.

\paragraph{Consistency.}
Following the consistency proofs of \citet{shekhar2023nonparametric} and \citet[Theorem~2.4]{podkopaev2023sequential}, suppose \(\|\Delta\| > \varepsilon\) at some point.
The running mean \(D_{r-1}\) converges to \(\Delta\) by the strong law of large numbers, so the witness converges to \(\Delta/\|\Delta\|\) and the conditional mean of the bets to \(\|\Delta\|/(2\sqrt{\kappa}) > 0\).
The log-wealth at that point then grows at the positive rate \(g^\star\) of \Cref{app:severity-detectability}, \(\log \bar E_r\) grows at least as fast by the bound above, and the test rejects with probability one.

\subsection{Severity and detectability}
\label{app:severity-detectability}

\paragraph{The oracle bet.}
Above the noise floor, the witness converges to \(u := \Delta/\|\Delta\|\).
With this witness fixed, the bets
\begin{equation*}
    b_r := \frac{\langle u, \hat\delta_r \rangle}{2\sqrt{\kappa}} \in [-1, 1]
\end{equation*}
are independent and identically distributed, with mean and variance
\begin{equation*}
    m := \frac{\|\Delta\|}{2\sqrt{\kappa}},
    \qquad
    s^2 := \frac{\Var\big(\langle u, \hat\delta_r \rangle\big)}{4\kappa},
\end{equation*}
where \(s^2\) includes any Monte Carlo error in \(\hat\delta_r\).
Write \(\rho := m^2/s^2\) for the squared signal-to-noise ratio along the bias.
With exact embeddings, the Cauchy--Schwarz inequality gives \(\Var(\langle u, \delta_r \rangle) \le \mathbb{E}\big[\|\delta_r - \Delta\|^2\big] = V\), so \(\rho \ge \Psi\), with equality only when the spread lies entirely along the bias.
By the logarithmic regret of the stakes (\Cref{app:testing}), the log-wealth grows asymptotically at rate
\begin{equation*}
    g^\star := \max_{\lambda \in [0, 1/2]} \mathbb{E}\big[\log(1 + \lambda b_r)\big],
\end{equation*}
which plays the role of power for such tests~\citep{grunwald2024safe}, and the test rejects after roughly \(\log(1/\alpha)/g^\star\) updates.

\paragraph{Two regimes.}
Up to constants, the bounds below are those of \citet[Theorem~2.4 and Proposition~2.5]{podkopaev2023sequential} in our notation, which \citet{shekhar2023nonparametric} also give for two-sample tests.
For \(|x| \le 1/2\),
\begin{equation*}
    x - x^2 \;\le\; \log(1+x) \;\le\; x - \tfrac{3}{8}x^2.
\end{equation*}
Apply these with \(x = \lambda b_r\), using \(\mathbb{E}[b_r^2] = m^2 + s^2\).
The lower bound gives \(\mathbb{E}[\log(1+\lambda b_r)] \ge \lambda m - \lambda^2(m^2+s^2)\).
The stake \(\lambda = m/(2(m^2+s^2))\) attains \(\tfrac{1}{4}\rho/(1+\rho)\) when it is at most \(1/2\), and otherwise \(\lambda = 1/2\) attains more than \(m/4\).
The upper bound gives \(\mathbb{E}[\log(1+\lambda b_r)] \le \lambda m - \tfrac{3}{8}\lambda^2(m^2+s^2)\), whose maximum over \(\lambda\) is \(\tfrac{2}{3}\rho/(1+\rho)\), and \(\log(1+x) \le x\) gives \(\mathbb{E}[\log(1+\lambda b_r)] \le m/2\).
Together,
\begin{align*}
    g^\star &\ge \tfrac{1}{4}\min\Big\{\frac{\rho}{1+\rho},\; m\Big\}, \\
    g^\star &\le \min\Big\{\tfrac{2}{3}\,\frac{\rho}{1+\rho},\; \frac{m}{2}\Big\}.
\end{align*}

The growth rate is thus governed, up to constants, by the smaller of \(\rho/(1+\rho)\), the systematic share along the bias, and \(m = \|\Delta\|/(2\sqrt{\kappa})\), the bias relative to the kernel bound.
The latter enters because bets are scaled by the worst-case bound \(\|\hat\delta_r\| \le 2\sqrt{\kappa}\), which keeps the wealth non-negative whatever the procedure does.
The stake cap then stops the test betting more heavily when bets are small.
A single update typically moves the embedding far less than \(2\sqrt{\kappa}\), so \(m\) is small and, unless \(\rho/(1+\rho) < m\), the rate lies between \(m/4\) and \(m/2\), scaling with the absolute bias and barely depending on the spread.
This is the low-variance regime of \citet{shekhar2023nonparametric}, in which the rate improves from order \(m^2\) to order \(m\), as \citet{podkopaev2023sequential} also show.
Since \(\Psi\) is scale invariant, procedures with the same \(\Psi\) but different absolute bias are rejected at different speeds, as for the incTNP and the fixed sparse GPs of similar \(\Psi\) in \Cref{fig:synthetic-gp}c.
With exact embeddings, \(\rho \ge \Psi\) gives \(g^\star \ge \tfrac{1}{4}\min\{\Psi/(1+\Psi),\, m\}\), so \(\Psi\) alone bounds the rate from below only when \(\Psi/(1+\Psi) \le m\), as in the low signal-to-noise regime.

\paragraph{Conservativeness.}
Scaling bets by an estimated displacement size would add power, but bets that could then exceed one in magnitude would need clipping, which breaks the mean-zero property under the null unless the displacements are symmetric.
For a non-negative kernel, such as the Dirac kernel, \(\|\hat\delta_r\|^2 \le 2\kappa\), so bets could be scaled by \(\sqrt{2\kappa}\) without loss of validity.
A tighter almost-sure bound on \(\hat\delta_r\) would need assumptions on how far an update can move the predictive, and exact embeddings, since Monte Carlo error alone can make \(\hat\delta_r\) as large as the kernel allows.
Asymptotic anytime-valid inference~\citep{waudbysmith2024time} avoids the range bound and adapts to the variance of the bets, but is valid only asymptotically.
We keep the non-asymptotic test because it is meant to audit exact coherence, where validity matters more than speed, and grade severity with \(\Psi\).

\section{Experimental details}
\label{app:experiments}

This appendix gives the details needed to reproduce each experiment in \Cref{sec:experiments}, one subsection per experiment, and the code is available at \url{https://github.com/jreaso/predictive-coherence}.

\paragraph{Compute.}
The sparse GP, tabular and Bayesian linear regression experiments used about 155 CPU core-hours and 4 GPU-hours on one laptop, and the neural testbed about 8 hours on one NVIDIA H100 GPU.
Preliminary runs used roughly as much again.

\subsection{Sparse GPs}
\label{app:sparse-gp}

The data are drawn from the GP prior, so the exact GP is the Bayesian ground truth for the sparse GPs and the incTNP.

\paragraph{Environments and tasks.}
Each of 20 environments is a function drawn from a GP prior (squared-exponential kernel, lengthscale 0.25, unit variance), observed with noise of standard deviation 0.1 at 512 inputs drawn uniformly from \([-2, 2]\).
Each environment is split at random 10 times into 24 context points, a block of 8 and 64 evaluation points.

\paragraph{Sparse GPs.}
Inducing points are a subset of the context inputs, chosen by greedy variance selection~\citep{burt2020convergence}, and \(q(u)\) is the optimum of \citet{titsias2009variational}.
Both update procedures start from this fit.
With fixed inducing points, the update keeps them and updates \(q(u)\), whereas re-selection chooses up to \(M\) afresh from all 32 inputs and refits.
The predictive includes the variance that the inducing points leave unexplained, but the update absorbs new data through the projected likelihood, which omits it.
This mismatch is the source of the incoherence.
\Cref{fig:synthetic-gp} and \Cref{tab:sgp-results} show \(M = 10, 12, \dots, 24\), although re-selection at \(M = 24\), with \(\Psi = 5 \times 10^{-6}\), lies below the figure's axes.
Near \(M = 24\), selection stops once the context is explained to numerical precision, so some tasks use fewer than \(M\) inducing points.

\paragraph{incTNP.}
The incTNP uses the sequential architecture of \citet{mortimer2026incremental}, which updates by extending a key--value cache rather than re-encoding the context.
It has five layers of width 128 and eight attention heads, and a Gaussian output for each target.
We train it on sequences of 49 points from the same prior, predicting each point from those before it, for 400\,000 steps with batches of 32.
Training uses Muon~\citep{jordan2024muon} on hidden weight matrices and AdamW~\citep{loshchilov2019decoupled} elsewhere, at learning rate \(5 \times 10^{-4}\) and weight decay 0.01, with 40\,000 warm-up steps followed by cosine decay.

\paragraph{Measuring and testing incoherence.}
Each round draws the block's outputs from the model's own joint predictive, autoregressively for the incTNP.
The kernel on outputs is Gaussian with lengthscale 0.1, the noise standard deviation, and is approximated by 256 random Fourier features.
Every predictive is Gaussian, so the embeddings are exact.
We run 1000 rounds for each task and model, shared by the estimator and the test of \Cref{app:testing}.
For the exact and sparse GPs, \(\Psi\) is also available in closed form, which \Cref{fig:synthetic-gp} and \Cref{tab:sgp-results} report.

\paragraph{Accuracy.}
Accuracy is the KL divergence from the exact GP's predictive to the model's, computed between their marginals at each evaluation point and averaged over the points.
After the update, both are conditioned on the block's observed outputs.
\Cref{fig:synthetic-gp} reports its geometric mean over tasks.
\(\Psi\) is the same in panels (a) and (b), since it describes a single update of the fitted model.

\paragraph{Results.}
\Cref{tab:sgp-results} gives the values behind \Cref{fig:synthetic-gp}, and \Cref{tab:sgp-rounds} the estimates of \(\Psi\) and their leave-one-round-out standard errors after 10, 100 and 1000 rounds.
The estimates agree with the closed form within their standard errors, and match it from 100 rounds.
Between 100 and 1000 rounds, the standard error falls roughly as \(R^{-1/2}\) for clearly incoherent models and roughly as \(R^{-1}\) near coherence, where the U-statistic is degenerate.

\begin{table}[t]
\centering
\caption{Results of the sparse GP experiment over 200 tasks.
\(\Psi\) is the mean over tasks with one standard error over environments, in closed form for the GP-based models and estimated from 1000 rounds for the incTNP.
KL is the geometric mean over tasks, in nats per evaluation point, before and after the update, with values below \(10^{-10}\) clamped before averaging.
Rejected is the share of tasks rejected within 1000 rounds at level 0.05.}
\label{tab:sgp-results}
\small
\begin{tabular}{@{}l r l l l r@{}}
\toprule
Model & \(M\) & \(\Psi\) & KL before & KL after & Rejected \\
\midrule
Exact GP & -- & \(0\) & \(0\) & \(0\) & \(0.00\) \\
incTNP & -- & \(0.0130 \pm 0.0008\) & \(3.1 \times 10^{-3}\) & \(2.9 \times 10^{-3}\) & \(0.99\) \\
\addlinespace
Fixed & \(10\) & \(0.124 \pm 0.006\) & \(0.47\) & \(0.66\) & \(0.99\) \\
& \(12\) & \(0.093 \pm 0.005\) & \(0.18\) & \(0.35\) & \(0.99\) \\
& \(14\) & \(0.066 \pm 0.005\) & \(4.3 \times 10^{-2}\) & \(0.17\) & \(0.95\) \\
& \(16\) & \(0.040 \pm 0.004\) & \(4.5 \times 10^{-3}\) & \(7.0 \times 10^{-2}\) & \(0.83\) \\
& \(18\) & \(0.026 \pm 0.003\) & \(1.6 \times 10^{-4}\) & \(3.1 \times 10^{-2}\) & \(0.66\) \\
& \(20\) & \(0.016 \pm 0.003\) & \(2.1 \times 10^{-6}\) & \(1.4 \times 10^{-2}\) & \(0.46\) \\
& \(22\) & \(0.013 \pm 0.002\) & \(6.0 \times 10^{-9}\) & \(7.1 \times 10^{-3}\) & \(0.40\) \\
& \(24\) & \(0.012 \pm 0.002\) & \(3.5 \times 10^{-10}\) & \(5.3 \times 10^{-3}\) & \(0.38\) \\
\addlinespace
Re-selected & \(10\) & \(1.16 \pm 0.08\) & \(0.47\) & \(0.66\) & \(1.00\) \\
& \(12\) & \(0.93 \pm 0.08\) & \(0.18\) & \(0.34\) & \(1.00\) \\
& \(14\) & \(0.45 \pm 0.04\) & \(4.3 \times 10^{-2}\) & \(0.12\) & \(0.99\) \\
& \(16\) & \(0.153 \pm 0.021\) & \(4.5 \times 10^{-3}\) & \(2.9 \times 10^{-2}\) & \(0.99\) \\
& \(18\) & \(0.021 \pm 0.003\) & \(1.6 \times 10^{-4}\) & \(3.3 \times 10^{-3}\) & \(0.83\) \\
& \(20\) & \(0.0022 \pm 0.0003\) & \(2.1 \times 10^{-6}\) & \(2.3 \times 10^{-4}\) & \(0.29\) \\
& \(22\) & \((1.4 \pm 0.3) \times 10^{-4}\) & \(6.0 \times 10^{-9}\) & \(9.6 \times 10^{-6}\) & \(0.02\) \\
& \(24\) & \(5 \times 10^{-6}\) & \(3.5 \times 10^{-10}\) & \(1.9 \times 10^{-7}\) & \(0.00\) \\
\bottomrule
\end{tabular}
\end{table}

\begin{table}[t]
\centering
\caption{Estimates of \(\Psi\) after \(R\) rounds, as the mean over tasks \(\pm\) the root mean square of the per-task leave-one-round-out standard errors.
The estimates for different \(R\) are prefixes of the same rounds.}
\label{tab:sgp-rounds}
\small
\begin{tabular}{@{}l l l l l@{}}
\toprule
Model & Closed form & \(R = 10\) & \(R = 100\) & \(R = 1000\) \\
\midrule
Exact GP & \(0\) & \(-0.005 \pm 0.090\) & \(0.000 \pm 0.008\) & \(0.0000 \pm 0.0007\) \\
incTNP & -- & \(0.017 \pm 0.093\) & \(0.014 \pm 0.010\) & \(0.013 \pm 0.002\) \\
Fixed, \(M = 10\) & \(0.124\) & \(0.127 \pm 0.141\) & \(0.123 \pm 0.019\) & \(0.124 \pm 0.005\) \\
Fixed, \(M = 24\) & \(0.012\) & \(0.016 \pm 0.111\) & \(0.012 \pm 0.011\) & \(0.012 \pm 0.002\) \\
Re-selected, \(M = 20\) & \(0.0022\) & \(0.001 \pm 0.093\) & \(0.003 \pm 0.008\) & \(0.0022 \pm 0.0008\) \\
\bottomrule
\end{tabular}
\end{table}

\subsection{Tabular data}
\label{app:tabular}

\paragraph{Data and splits.}
We use three regression and five classification datasets from OpenML (\Cref{tab:tabular-datasets}).
The seven UCI datasets are under CC BY 4.0, and california is listed as public on OpenML.
We drop rows with missing values, one-hot encode categorical inputs, and standardise inputs and regression outputs with the training set's mean and standard deviation.
A split divides a dataset at random into a training set (half the rows, at most 1000), a test set (15\%, at most 100) and a pool (the rest, at most 2000).
Each round draws block inputs, a quarter as many as the training set, without replacement from the pool.
The design is therefore random, and \(Q_Y\) and \(V\) average over it as well as over the block's outputs.
Coherence for every design implies coherence in this sense, but not conversely.
Every model is fitted on the training set and updated on the training set and block.
All models use the same ten splits per dataset, block inputs and 500 rounds per split, except TabPFN, which because of its cost is run on the first split only, for 128 rounds.

\begin{table}[t]
\centering
\caption{Datasets and the sizes of each part of a split.
Inputs counts features after one-hot encoding.
Sources: concrete~\citep{uci_concrete,yeh1998modeling}, california~\citep{pace1997sparse}, yacht~\citep{uci_yacht}, australian~\citep{uci_australian}, cleveland~\citep{uci_heart_disease,detrano1989international}, german~\citep{uci_german_credit}, vehicle~\citep{uci_vehicle} and wdbc~\citep{uci_wdbc}, all fetched from OpenML~\citep{vanschoren2013openml}.}
\label{tab:tabular-datasets}
\small
\begin{tabular}{@{}l r r r r r r r r@{}}
\toprule
Dataset & OpenML id & Rows & Inputs & Classes & Training & Test & Pool & Block \\
\midrule
\multicolumn{9}{@{}l}{\emph{Regression}} \\
concrete & \(44959\) & \(1030\) & \(8\) & -- & \(515\) & \(100\) & \(415\) & \(128\) \\
california & \(44977\) & \(20640\) & \(8\) & -- & \(1000\) & \(100\) & \(2000\) & \(250\) \\
yacht & \(42370\) & \(308\) & \(6\) & -- & \(154\) & \(46\) & \(108\) & \(38\) \\
\addlinespace
\multicolumn{9}{@{}l}{\emph{Classification}} \\
australian & \(40981\) & \(690\) & \(42\) & \(2\) & \(345\) & \(100\) & \(245\) & \(86\) \\
cleveland & \(194\) & \(297\) & \(25\) & \(5\) & \(148\) & \(44\) & \(105\) & \(37\) \\
german & \(31\) & \(1000\) & \(61\) & \(2\) & \(500\) & \(100\) & \(400\) & \(125\) \\
vehicle & \(54\) & \(846\) & \(18\) & \(4\) & \(423\) & \(100\) & \(323\) & \(105\) \\
wdbc & \(1510\) & \(569\) & \(30\) & \(2\) & \(284\) & \(85\) & \(200\) & \(71\) \\
\bottomrule
\end{tabular}
\end{table}

\paragraph{Exact models.}
The GP (ARD squared-exponential kernel, hyperparameters maximising the marginal likelihood) conditions on the block.
Exact logistic regression, a softmax model with a Gaussian prior on its weights, represents its posterior by 8192 HMC draws~\citep{neal2011mcmc} and updates by reweighting them, which is exact Bayes with the draws as a discrete prior.

\paragraph{Approximately Bayesian models.}
These recompute their approximation to update.
The sparse GP takes the GP's hyperparameters and the optimal \(q(u)\) of \citet{titsias2009variational}, and updates \(q(u)\) with its inducing points fixed (\Cref{app:sparse-gp}).
Greedy variance selection picks these from the training inputs until the unexplained prior variance per input falls below the noise variance, giving 29 to 100 points.
The refitted GP refits the GP's hyperparameters on the training set and block before conditioning.
The full and diagonal Laplace approximations~\citep{mackay1992evidence} share exact logistic regression's prior and move to the new mode, taking the full precision or its diagonal there.
BART~\citep{chipman2010bart} and XBART~\citep{he2019xbart} (\texttt{stochtree} 0.4.5~\citep{herren2025stochtree}, default priors, 200 trees) predict with the mixture over 100 MCMC draws from one chain (BART) or the last 25 of 40 grow-from-root sweeps (XBART), and update by rerunning the sampler with a fresh seed.

\paragraph{Plug-in models.}
These refit to update.
For regression, kernel ridge and nearest neighbours predict a Gaussian at the fitted function or at the mean of the \(k\) nearest outputs, with the mean squared leave-one-out residual as variance.
For classification, plug-in logistic regression predicts at the posterior mode of exact logistic regression, and nearest neighbours the class frequencies among the \(k\) neighbours.
CatBoost~\citep{prokhorenkova2018catboost} predicts a mean and variance (\texttt{RMSEWithUncertainty} loss) or class probabilities, with its seed fixed at zero.
A plug-in model's joint is by default the product of its predictives.
Kernel ridge, plug-in logistic regression and nearest neighbours are also run with the prequential joint of \Cref{sec:coherence}, in random order, which leaves the current predictive, and so its accuracy, unchanged.

\paragraph{TabPFN.}
TabPFN~\citep{hollmann2025tabpfn} (v2.5 weights~\citep{grinsztajn2025tabpfn}, TabPFN-2.5 licence, one estimator, softmax temperature one, not the default 0.9) conditions on the training set and updates by appending the block to its context.
It draws the block's outputs autoregressively, in the block's order.
The network variant switches off the package's input and target transforms, fingerprint feature, and feature and class shifts, keeping only the clipping of outlying inputs for classification.
The preprocessing variant uses the default configuration, whose transforms are refitted whenever the context changes.

\paragraph{Hyperparameters.}
Kernel ridge's penalty and bandwidth, the number of neighbours, CatBoost's depth and number of trees, and the standard deviation of the logistic regression prior are chosen once per dataset by five-fold cross-validated log likelihood on the first split's training set.
This set contains about half of each other split's test set, except on california.
A regression dataset's noise scale is the smallest cross-validated RMSE among the tuned models and the GP, BART and XBART, which are scored only for this, and is 0.309 on concrete, 0.499 on california and 0.042 on yacht.

\paragraph{Evaluation.}
Regression uses a Gaussian kernel with the noise scale as lengthscale, approximated by 256 random Fourier features~\citep{rahimi2007random} shared by every model, and classification the Dirac kernel, both giving every predictive an exact embedding.
The evaluation points are the test inputs, and the test is that of \Cref{app:testing}, sharing the rounds of the \(\Psi\) estimate.
Accuracy is the current predictive's test log loss, with class probabilities clamped below at \(10^{-16}\), which matters only for nearest neighbours, whose predictive can give a label probability zero.

\paragraph{Results.}
\Cref{tab:tabular-summary} summarises the results over datasets, and \Cref{tab:tabular-psi,tab:tabular-logloss} give \(\Psi\) and test log loss on each.
The exact models have \(|\Psi| < 5 \times 10^{-4}\) on every dataset.
Exact logistic regression on cleveland is 4.5 standard errors below zero because near coherence the estimate is right-skewed (\Cref{app:ratio-estimation}), so most splits fall below its mean, and rerunning one split 60 times gives an estimate centred on zero.
For the inexact models, the standard deviation of \(\Psi\) over the ten splits is 1.8 to 15 times its within-split standard error (median over datasets), so TabPFN's standard errors, from its one split, understate its variation across splits.
Across the models of \Cref{tab:tabular-summary}, the Spearman correlation of \(\Psi\) between datasets is 0.98 for every regression pair and at least 0.88 for classification, lowest for australian and german.

\paragraph{Robustness.}
Halving or doubling the lengthscale reorders models only on yacht, swapping BART and CatBoost, XBART and nearest neighbours, and the TabPFN variants, each pair within 15\% (\Cref{tab:tabular-lengthscales}).
A lengthscale of one, 2 to 24 times the noise scale, shrinks small values of \(\Psi\) and swaps models up to 23\% apart.
The prequential joints also change \(\Psi\), for example from 0.12 to 0.19 for kernel ridge on yacht.

\paragraph{Monte Carlo error in BART.}
BART's incoherence is largely Monte Carlo error.
On the first split of each regression dataset, rerunning the sampler with fresh seeds on the training set alone gives a squared discrepancy \(\|\Delta\|^2\) which is 1.1 to 1.2 times that of an update.
A run's 100 draws are worth about two independent draws on concrete and california, and fewer than one on yacht.

\paragraph{Detectability.}
Small incoherence often goes unrejected within the budget.
The sparse GP's \(\Psi\) exceeds twice its standard error on all 30 splits, but only 4 are rejected within 500 rounds, against a predicted median of about 1500.
The predicted round, \(\min_l \{1/\hat\Psi_l + 4\log(L/\alpha)/\|\hat\Delta_l\|\}\) over test inputs \(l\), is a heuristic for the rounds the witness needs to align plus those needed to reach \(\log(L/\alpha)\) at the upper bound \(m/2\) on the growth rate (\Cref{app:severity-detectability}), so it is optimistic.
TabPFN's predicted rounds, 127 to 585, mostly exceed its budget of 128.

\begin{table}[t]
\centering
\caption{Summary over datasets.
\(\Psi\) is the median over datasets of the model's mean over splits, with its range.
Rank is the median over datasets of the model's rank by test log loss among the models listed for its task (1 is best, ties averaged).
Rej.\ is the share of the 30 (regression) or 50 (classification) splits rejected within 500 rounds.
TabPFN, never rejected on its one split within 128 rounds, is shown as -- since this is not comparable.}
\label{tab:tabular-summary}
\footnotesize
\setlength{\tabcolsep}{3pt}
\begin{tabular}[t]{@{}l r l r r@{}}
\toprule
\multicolumn{5}{@{}l}{\emph{Regression}} \\
Model & \(\Psi\) & Range & Rank & Rej. \\
\midrule
GP & \(0.000\) & \(0.000\)--\(0.000\) & \(3.5\) & \(0.00\) \\
\addlinespace
Sparse GP & \(0.007\) & \(0.004\)--\(0.018\) & \(9\) & \(0.13\) \\
GP, refitted & \(0.002\) & \(0.001\)--\(0.002\) & \(3.5\) & \(0.03\) \\
BART & \(0.710\) & \(0.632\)--\(0.863\) & \(7\) & \(1.00\) \\
XBART & \(0.378\) & \(0.271\)--\(0.787\) & \(5\) & \(1.00\) \\
\addlinespace
Kernel ridge & \(0.013\) & \(0.008\)--\(0.119\) & \(8\) & \(0.37\) \\
Nearest neighbours & \(0.470\) & \(0.461\)--\(0.701\) & \(10\) & \(1.00\) \\
CatBoost & \(0.900\) & \(0.596\)--\(0.912\) & \(7\) & \(1.00\) \\
\addlinespace
TabPFN, network & \(0.102\) & \(0.089\)--\(0.195\) & \(2\) & -- \\
TabPFN, preproc. & \(0.118\) & \(0.083\)--\(0.222\) & \(1\) & -- \\
\bottomrule
\end{tabular}
\hfill
\begin{tabular}[t]{@{}l r l r r@{}}
\toprule
\multicolumn{5}{@{}l}{\emph{Classification}} \\
Model & \(\Psi\) & Range & Rank & Rej. \\
\midrule
Logistic, exact & \(0.000\) & \(0.000\)--\(0.000\) & \(4\) & \(0.00\) \\
\addlinespace
Full Laplace & \(0.004\) & \(0.001\)--\(0.007\) & \(4\) & \(0.02\) \\
Diagonal Laplace & \(0.008\) & \(0.002\)--\(0.012\) & \(5\) & \(0.00\) \\
\addlinespace
Logistic, plug-in & \(0.007\) & \(0.001\)--\(0.016\) & \(3\) & \(0.10\) \\
Nearest neighbours & \(0.883\) & \(0.734\)--\(0.927\) & \(8\) & \(1.00\) \\
CatBoost & \(0.273\) & \(0.145\)--\(0.339\) & \(6\) & \(1.00\) \\
\addlinespace
TabPFN, network & \(0.071\) & \(0.046\)--\(0.190\) & \(2\) & -- \\
TabPFN, preproc. & \(0.126\) & \(0.054\)--\(0.425\) & \(2\) & -- \\
\bottomrule
\end{tabular}
\end{table}

\begin{table}[t]
\centering
\caption{\(\Psi\) by dataset: the mean over ten splits \(\pm\) their standard deviation divided by \(\sqrt{10}\), or for TabPFN its one split \(\pm\) the leave-one-round-out standard error.}
\label{tab:tabular-psi}
\footnotesize
\setlength{\tabcolsep}{4pt}
\begin{tabular}{@{}l c c c@{}}
\toprule
Model & concrete & california & yacht \\
\midrule
GP (\(\times 10^{-4}\)) & \(-0.6 \pm 1.4\) & \(2.3 \pm 2.2\) & \(2.2 \pm 2.4\) \\
\addlinespace
Sparse GP & \(0.007 \pm 0.001\) & \(0.004 \pm 0.000\) & \(0.018 \pm 0.004\) \\
GP, refitted & \(0.002 \pm 0.001\) & \(0.001 \pm 0.000\) & \(0.002 \pm 0.001\) \\
BART & \(0.632 \pm 0.035\) & \(0.710 \pm 0.051\) & \(0.863 \pm 0.073\) \\
XBART & \(0.378 \pm 0.034\) & \(0.271 \pm 0.018\) & \(0.787 \pm 0.075\) \\
\addlinespace
Kernel ridge & \(0.013 \pm 0.002\) & \(0.008 \pm 0.001\) & \(0.119 \pm 0.012\) \\
Nearest neighbours & \(0.461 \pm 0.020\) & \(0.470 \pm 0.017\) & \(0.701 \pm 0.067\) \\
CatBoost & \(0.900 \pm 0.048\) & \(0.596 \pm 0.045\) & \(0.912 \pm 0.239\) \\
\addlinespace
TabPFN, network & \(0.089 \pm 0.008\) & \(0.102 \pm 0.008\) & \(0.195 \pm 0.019\) \\
TabPFN, preproc. & \(0.083 \pm 0.006\) & \(0.118 \pm 0.009\) & \(0.222 \pm 0.018\) \\
\bottomrule
\end{tabular}

\vspace{0.8em}
\begin{tabular}{@{}l c c c c c@{}}
\toprule
Model & australian & cleveland & german & vehicle & wdbc \\
\midrule
Logistic, exact (\(\times 10^{-4}\)) & \(2.3 \pm 2.0\) & \(-4.5 \pm 1.0\) & \(0.0 \pm 1.7\) & \(0.9 \pm 2.4\) & \(-1.2 \pm 2.0\) \\
\addlinespace
Full Laplace & \(0.003 \pm 0.001\) & \(0.004 \pm 0.001\) & \(0.001 \pm 0.000\) & \(0.007 \pm 0.002\) & \(0.004 \pm 0.002\) \\
Diagonal Laplace & \(0.008 \pm 0.001\) & \(0.006 \pm 0.001\) & \(0.002 \pm 0.000\) & \(0.009 \pm 0.001\) & \(0.012 \pm 0.002\) \\
\addlinespace
Logistic, plug-in & \(0.007 \pm 0.003\) & \(0.003 \pm 0.001\) & \(0.001 \pm 0.000\) & \(0.012 \pm 0.008\) & \(0.016 \pm 0.008\) \\
Nearest neighbours & \(0.897 \pm 0.038\) & \(0.849 \pm 0.036\) & \(0.734 \pm 0.042\) & \(0.883 \pm 0.026\) & \(0.927 \pm 0.059\) \\
CatBoost & \(0.145 \pm 0.011\) & \(0.339 \pm 0.023\) & \(0.211 \pm 0.012\) & \(0.327 \pm 0.012\) & \(0.273 \pm 0.043\) \\
\addlinespace
TabPFN, network & \(0.190 \pm 0.034\) & \(0.046 \pm 0.010\) & \(0.071 \pm 0.013\) & \(0.142 \pm 0.020\) & \(0.048 \pm 0.013\) \\
TabPFN, preproc. & \(0.098 \pm 0.012\) & \(0.054 \pm 0.008\) & \(0.425 \pm 0.034\) & \(0.248 \pm 0.022\) & \(0.126 \pm 0.025\) \\
\bottomrule
\end{tabular}
\end{table}

\begin{table}[t]
\centering
\caption{Test log loss in nats per output, on the standardised outputs for regression: the mean over ten splits, or TabPFN's one split.
The lowest on each dataset before rounding is in bold.}
\label{tab:tabular-logloss}
\footnotesize
\begin{tabular}{@{}l r r r@{}}
\toprule
Model & concrete & california & yacht \\
\midrule
GP & \(0.283\) & \(0.738\) & \(-1.803\) \\
\addlinespace
Sparse GP & \(0.457\) & \(0.889\) & \(-1.749\) \\
GP, refitted & \(0.283\) & \(0.738\) & \(-1.803\) \\
BART & \(0.151\) & \(0.773\) & \(-1.540\) \\
XBART & \(0.089\) & \(0.738\) & \(-0.128\) \\
\addlinespace
Kernel ridge & \(0.412\) & \(0.762\) & \(-0.231\) \\
Nearest neighbours & \(0.892\) & \(0.956\) & \(1.045\) \\
CatBoost & \(0.400\) & \(0.877\) & \(-1.601\) \\
\addlinespace
TabPFN, network & \(-0.254\) & \(0.291\) & \(\mathbf{-3.021}\) \\
TabPFN, preproc. & \(\mathbf{-0.308}\) & \(\mathbf{0.237}\) & \(-2.959\) \\
\bottomrule
\end{tabular}

\vspace{0.8em}
\begin{tabular}{@{}l r r r r r@{}}
\toprule
Model & australian & cleveland & german & vehicle & wdbc \\
\midrule
Logistic, exact & \(0.355\) & \(1.056\) & \(0.488\) & \(0.434\) & \(0.088\) \\
\addlinespace
Full Laplace & \(0.352\) & \(1.043\) & \(0.490\) & \(0.449\) & \(0.094\) \\
Diagonal Laplace & \(0.356\) & \(1.043\) & \(0.493\) & \(0.443\) & \(0.107\) \\
\addlinespace
Logistic, plug-in & \(0.356\) & \(1.074\) & \(\mathbf{0.488}\) & \(0.428\) & \(0.086\) \\
Nearest neighbours & \(0.409\) & \(1.421\) & \(0.598\) & \(0.840\) & \(0.225\) \\
CatBoost & \(0.323\) & \(1.190\) & \(0.490\) & \(0.500\) & \(0.097\) \\
\addlinespace
TabPFN, network & \(\mathbf{0.211}\) & \(0.946\) & \(0.497\) & \(\mathbf{0.340}\) & \(0.049\) \\
TabPFN, preproc. & \(0.218\) & \(\mathbf{0.916}\) & \(0.512\) & \(0.348\) & \(\mathbf{0.034}\) \\
\bottomrule
\end{tabular}
\end{table}

\begin{table}[t]
\centering
\caption{\(\Psi\) with the kernel lengthscale at the noise scale / half it / twice it / one (the training outputs' standard deviation).
The GP, with \(|\Psi| < 5 \times 10^{-4}\) at every lengthscale, is omitted.}
\label{tab:tabular-lengthscales}
\footnotesize
\setlength{\tabcolsep}{4pt}
\begin{tabular}{@{}l c c c@{}}
\toprule
Model & concrete & california & yacht \\
\midrule
Sparse GP & \(0.007\) / \(0.008\) / \(0.005\) / \(0.003\) & \(0.004\) / \(0.004\) / \(0.002\) / \(0.002\) & \(0.018\) / \(0.020\) / \(0.015\) / \(0.001\) \\
GP, refitted & \(0.002\) / \(0.002\) / \(0.002\) / \(0.002\) & \(0.001\) / \(0.001\) / \(0.001\) / \(0.001\) & \(0.002\) / \(0.002\) / \(0.001\) / \(0.001\) \\
BART & \(0.632\) / \(0.641\) / \(0.617\) / \(0.610\) & \(0.710\) / \(0.710\) / \(0.708\) / \(0.708\) & \(0.863\) / \(0.941\) / \(0.783\) / \(0.674\) \\
XBART & \(0.378\) / \(0.389\) / \(0.364\) / \(0.357\) & \(0.271\) / \(0.277\) / \(0.261\) / \(0.261\) & \(0.787\) / \(0.860\) / \(0.699\) / \(0.540\) \\
\addlinespace
Kernel ridge & \(0.013\) / \(0.015\) / \(0.008\) / \(0.005\) & \(0.008\) / \(0.010\) / \(0.005\) / \(0.005\) & \(0.119\) / \(0.118\) / \(0.112\) / \(0.020\) \\
Nearest neighbours & \(0.461\) / \(0.468\) / \(0.452\) / \(0.441\) & \(0.470\) / \(0.473\) / \(0.463\) / \(0.463\) & \(0.701\) / \(0.704\) / \(0.703\) / \(0.683\) \\
CatBoost & \(0.900\) / \(0.913\) / \(0.876\) / \(0.860\) & \(0.596\) / \(0.601\) / \(0.591\) / \(0.591\) & \(0.912\) / \(0.920\) / \(0.840\) / \(0.823\) \\
\addlinespace
TabPFN, network & \(0.089\) / \(0.095\) / \(0.084\) / \(0.080\) & \(0.102\) / \(0.099\) / \(0.105\) / \(0.105\) & \(0.195\) / \(0.167\) / \(0.246\) / \(0.269\) \\
TabPFN, preproc. & \(0.083\) / \(0.092\) / \(0.078\) / \(0.075\) & \(0.118\) / \(0.121\) / \(0.115\) / \(0.115\) & \(0.222\) / \(0.274\) / \(0.192\) / \(0.153\) \\
\bottomrule
\end{tabular}
\end{table}

\subsection{Mean-field Bayesian linear regression}
\label{app:blr}

\paragraph{The two-parameter example (\Cref{fig:bed-objective-splitting}).}
The model is \(y = w_0 + w_1 x\) plus unit-variance Gaussian noise, with prior \(w \sim \mathcal{N}(0, 4I)\), fitted to five inputs drawn from \(U[4, 6]\) with outputs \(0.2x\) plus standard Gaussian noise.
The exact posterior correlation between intercept and slope is \(-0.97\), and the mean-field approximation is formed as in the experiment below.
The target is \(x_* = 3.5\), and the candidate design ranges over \([-10, 30]\) in steps of 0.1, of which the figure shows \([0, 18]\) and marks \(A = 2\) and \(B = 16\).
Under exact conditioning, EPIG and EMI both prefer \(A\) (0.081 against 0.068 nats).
Under mean-field VI, \(\EPIG_H\) prefers \(B\) (0.037 against 0.013), whereas EMI prefers \(A\) (0.045 against 0.014), as does \(\EPIG_{\KL}\) (0.046 against 0.015), which stays within 0.0012 nats of EMI over the range shown.

\paragraph{Environments.}
The acquisition experiment is entirely closed form, so there is no estimation noise beyond the choice of environments.
The model is Bayesian linear regression on 100 random Fourier features~\citep{rahimi2007random} of a Gaussian kernel with lengthscale 0.1 on a scalar input, with weights \(w \sim \mathcal{N}(0, 3I)\) and noise variance 0.8.
The truth is drawn from this prior, so the model is well specified.
Each of 100 environments, shared by every arm, draws its features, the true weights, 25 initial inputs from \(U[-1, 1]\), a pool of 1000 inputs and 100 target inputs from \(U[-10, 10]\), noisy labels for the initial and pool inputs, and a random order of the pool.
The reported log loss is the learner's expected log loss for a noisy observation from the truth at each target input, so there is no test set.

\paragraph{Learner and acquisition.}
The learner keeps the exact conjugate posterior.
Its mean-field approximation, the optimum of \(\KL(q \,\|\, p)\) over factorised Gaussians, takes the exact posterior means and the inverse diagonal of the exact posterior precision as variances.
It is refitted after each acquisition rather than updated in a streaming fashion.
Each arm acquires 100 points one at a time, maximising its score over the remaining pool.
The refit's variance does not depend on the outcome and its mean is affine in it, so \(Q_Y\) is Gaussian and \(\EPIG_H\) and EMI, averaged uniformly over the targets, are closed form.
\emph{Exact Bayes} acquires with the exact posterior, and random acquisition takes the pool in its random order.

\paragraph{The coherentised predictive.}
Rung \(\rho\) of the ladder scores candidates by \(\EPIG_H\) under a predictive made coherent for the mean-field refit on a reference set \(\mathcal{S}\), the first \(\round(\rho N)\) of the \(N\) remaining pool points in the environment's fixed order.
Coherentising an objective (\Cref{sec:cost}) would instead substitute \(Q_Y^a\) for \(P_Y\) at each candidate \(a\).
Every predictive here is \(\mathcal{N}(\phi(x)^\top \mu, \phi(x)^\top \Omega \phi(x) + \sigma^2)\) at \(x\), where \(\mu\) is the exact posterior mean and only the weight covariance \(\Omega\) varies.
With \(\Sigma_+\) and \(\Omega_+\) the exact and mean-field covariances after observing \(\mathcal{S}\), \(B = \sigma^{-2} \Sigma_+ \Phi_{\mathcal{S}}^\top\) and \(G = B \Phi_{\mathcal{S}}\), coherence requires
\begin{equation*}
    \Omega = \Omega_+ + \sigma^2 B B^\top + G \Omega G^\top .
\end{equation*}
As \(G\) has spectral radius below one, this discrete Lyapunov equation has the unique solution \(\sum_n G^n (\Omega_+ + \sigma^2 B B^\top) (G^\top)^n\), which we sum by repeated doubling.
In \(\EPIG_H(c)\), the coherentised predictive supplies \(H[P_Y]\) and the law of the outcome \(z\) at \(c\).
The post-update predictive \(P_Y^{c,z}\) is the mean-field refit on \(\mathcal{S} \cup \{c\}\), averaged over outcomes at \(\mathcal{S}\) drawn from the coherentised predictive given \(z\).
A candidate inside \(\mathcal{S}\) is therefore counted both as observed and as part of the block.
The reference set is recomputed at every step as the pool shrinks.
The shares are \(\rho \in \{0, 1, 2.5, 5, 10, 100\}\%\), and \(\rho = 0\) gives mean-field \(\EPIG_H\).

\paragraph{Measuring incoherence.}
\(\Psi\) is computed before any acquisition and pooled over the 100 targets, with a Gaussian kernel of lengthscale \(\sqrt{0.8}\), the noise standard deviation.
For each rung, the update is the mean-field refit on the whole pool, with outcomes drawn from the rung's predictive.
We use this block, larger than \Cref{sec:incoherence-ratio} would call for, because the ladder varies coherence with refits on shares of the pool.
\Cref{fig:blr}b plots its mean over environments.
For the mean-field model, the pool-scale \(\Psi\) is \(1.05 \times 10^{-3}\), whereas its \(\Psi\) for an update on a single candidate, the block the acquisition faces, is 0.046, pooled over candidates within each environment and averaged over environments.

\paragraph{Results.}
\Cref{tab:blr-results} gives the values behind \Cref{fig:blr}.
Every adjacent pair of rungs differs in loss by at least four standard errors of the paired difference over environments.
Coherentising over 1\%, 2.5\%, 5\% and 10\% of the pool recovers 32\%, 51\%, 80\% and 95\% of mean-field \(\EPIG_H\)'s excess loss.
Mean-field \(\EPIG_H\) and EMI choose different points at the first step in every environment, and their 100 acquired points overlap by 13 on average.

\begin{table}[t]
\centering
\caption{Results of the Bayesian linear regression experiment over 100 environments.
Excess loss is the mean-field learner's log loss less that after acquiring with the exact predictive, averaged over the 100 steps, with one standard error over environments.
\(\Psi\) is measured before acquisition for an update on the whole pool.}
\label{tab:blr-results}
\small
\begin{tabular}{@{}l l l@{}}
\toprule
Acquisition & Excess loss (nats) & \(\Psi\) (\(\times 10^{-3}\)) \\
\midrule
Mean-field, \(\EPIG_H\) (\(\rho = 0\)) & \(0.270 \pm 0.011\) & \(1.05 \pm 0.05\) \\
Coherentised, \(\rho = 1\%\) & \(0.184 \pm 0.009\) & \(0.80 \pm 0.04\) \\
Coherentised, \(\rho = 2.5\%\) & \(0.133 \pm 0.008\) & \(0.55 \pm 0.03\) \\
Coherentised, \(\rho = 5\%\) & \(0.054 \pm 0.005\) & \(0.32 \pm 0.02\) \\
Coherentised, \(\rho = 10\%\) & \(0.014 \pm 0.003\) & \(0.13 \pm 0.01\) \\
Coherentised, \(\rho = 100\%\) & \(0.002 \pm 0.002\) & \(0\) \\
\addlinespace
Mean-field, EMI & \(0.008 \pm 0.002\) & -- \\
Random & \(0.156 \pm 0.008\) & -- \\
\bottomrule
\end{tabular}
\end{table}

\paragraph{Robustness.}
We rerun the experiment with noise variance 0.1 or 3, with targets from \(U[4, 8]\), away from most of the pool, and with a truth on independent random features, so that the model is misspecified (\Cref{tab:blr-robustness}).
Mean-field \(\EPIG_H\) acquires worse data than random in every setting.
With low noise the mean-field variances are badly wrong: before acquisition, the mean-field log loss exceeds the exact one by 1.31 nats, against 0.045 in the baseline.
EMI, which still consumes these variances, then does no better than random, whereas the coherentised predictive, which changes them, still comes close to exact Bayes.

\begin{table}[t]
\centering
\caption{Excess loss as in \Cref{tab:blr-results}, with one standard error over environments, under changes to the baseline.
Low and high noise set the noise variance to 0.1 and 3, against 0.8 in the baseline.}
\label{tab:blr-robustness}
\small
\setlength{\tabcolsep}{4pt}
\begin{tabular}{@{}l l l l l l@{}}
\toprule
Setting & Mean-field \(\EPIG_H\) & Mean-field EMI & Random & \(\rho = 5\%\) & \(\rho = 100\%\) \\
\midrule
Baseline & \(0.270 \pm 0.011\) & \(0.008 \pm 0.002\) & \(0.156 \pm 0.008\) & \(0.054 \pm 0.005\) & \(0.002 \pm 0.002\) \\
Low noise & \(1.966 \pm 0.070\) & \(0.920 \pm 0.027\) & \(0.929 \pm 0.037\) & \(0.033 \pm 0.007\) & \(0.009 \pm 0.004\) \\
High noise & \(0.065 \pm 0.004\) & \(0.003 \pm 0.001\) & \(0.042 \pm 0.003\) & \(0.017 \pm 0.003\) & \(0.001 \pm 0.002\) \\
Distant targets & \(0.385 \pm 0.019\) & \(0.006 \pm 0.002\) & \(0.270 \pm 0.013\) & \(0.017 \pm 0.003\) & \(-0.001 \pm 0.002\) \\
Misspecified & \(0.537 \pm 0.018\) & \(0.016 \pm 0.003\) & \(0.454 \pm 0.016\) & \(0.204 \pm 0.009\) & \(0.008 \pm 0.002\) \\
\bottomrule
\end{tabular}
\end{table}

\subsection{Neural testbed}
\label{app:neural-testbed}

Our testbed is modelled on the Neural Testbed~\citep{osband2022neural}, with batch acquisition in place of its joint-prediction benchmark.

\paragraph{Environments.}
The truth is a random ReLU network \(f\) (two inputs, two hidden layers of 32 units, three outputs) with Xavier-normal weights~\citep{glorot2010understanding} and biases uniform on \(\pm 1/\sqrt{\text{fan-in}}\).
Its class probabilities are \(\softmax((f(x) - o)/0.35)\), with offsets \(o\) giving each class average probability \(1/3\) over a grid on \([-6, 6]^2\).
Inputs come from four fixed Gaussian clusters centred at \((\pm 3, \pm 3)\), each input's cluster uniform over those allowed.
The 24 training inputs use three clusters and standard deviation 0.5.
The pool of 512 uses all four, and the 64 targets and 4096 test inputs two, one with training data and one without, all with standard deviation 1.5.
Environments are kept if every class has mean true probability at least 0.1 over the test inputs, leaving 1000 of 1160.

\paragraph{The decision.}
Each environment has ten candidate batches of five pool inputs, shared by all models.
Eight are designed from per-input BALD (label--parameter mutual information) under a linearised diagonal Laplace approximation fitted to the training data.
They take the most informative inputs from the pool, target clusters or training clusters (greedily, with a minimum spacing), an informative input and its nearest neighbours, inputs spread over the pool or target clusters, inputs near the decision boundary, or a mix of target and training inputs.
The other two, and replacements for duplicated designed batches, are sampled with random cluster proportions, about three per environment.

\paragraph{Models.}
Network models share the truth's architecture without its temperature or offsets, and are fitted to the 24 training points.
Exact Bayes conditions 8192 networks drawn per environment from the generative prior, with temperature and offsets.
The full, Kronecker-factored~\citep{martens2015optimizing, ritter2018scalable} and diagonal linearised Laplace approximations~\citep{immer2021improving} share a MAP fit (Adam~\citep{kingma2015adam}, learning rate 0.05, 1000 steps, prior scale four times the generative one) and use generalised Gauss--Newton curvature~\citep{schraudolph2002fast}.
They condition the linearised model on the batch via a Laplace approximation in its logits (16 damped Newton steps).
Bayes by backprop~\citep{blundell2015weight} (Adam, learning rate 0.003, 8 samples) and IVON~\citep{shen2024variational} (learning rate 0.03, 4 samples) fit factorised Gaussians under the generative prior for 2000 steps, with KL weight 0.1.
They update by variational continual learning~\citep{nguyen2018variational}, with the current posterior as prior and the five new points as data (same KL weight, 100 steps, learning rates \(3 \times 10^{-4}\) and \(10^{-3}\) respectively).
The deep ensemble~\citep{lakshminarayanan2017simple} (8 members, Adam, learning rate 0.05, 1000 steps, weight penalty 3) and MC dropout~\citep{gal2016dropout} (rate 0.2, learning rate 0.05, 500 steps, weight penalty 1) update by continuing training on all 29 points, for 1000 and 50 steps respectively.
The Laplace and variational predictives mix 128 parameter draws, reparameterised by each posterior from standard-normal base draws fixed once per environment, and MC dropout's mixes 128 masks fixed once and shared by every predictive.
The incTNP~\citep{mortimer2026incremental} (eight layers of width 256, eight attention heads) updates by appending the batch to its context.
It is trained on generative tasks to predict each training point from its predecessors and each target from the training points and part of a batch, with Muon~\citep{jordan2024muon} on hidden weight matrices and AdamW~\citep{loshchilov2019decoupled} elsewhere (learning rate \(5 \times 10^{-4}\), cosine decay, 90\,000 steps of 128 tasks).
Hyperparameters were tuned on 64 separate environments by the KL from the truth before and after updating.
Only IVON's update learning rate and the ensemble's and MC dropout's update steps lie at the edge of a grid of over two values.

\paragraph{Objectives, incoherence and cost.}
For each candidate \(a\), the model is updated on all \(3^5 = 243\) labellings of the batch, so the objectives of \Cref{sec:objective-splitting} are exact sums weighted by its joint \(P_Z^a\), averaged over the 64 targets.
Mixtures label each input independently given a component, and the incTNP labels them autoregressively.
\(\Psi\) and \(\|\Delta\|^2\) use the Dirac kernel, pooled over the targets, and an environment's value is the median over its candidates' updates.
The random updates of Bayes by backprop, IVON and MC dropout are drawn twice from independent streams, and the objectives and losses use the first.
\(\|\Delta\|^2\) is estimated without bias by the inner product of the two draws' \(P_Z^a\)-weighted mean displacements, and \(V\) by their mean squared displacement less that estimate.
These estimates are negative in 1, 6 and 36 of the 1000 environments respectively, and are kept.
The distortion is the standard deviation of \({\EPIG_H} - {\EMI}\) over the ten candidates.
A candidate's loss is the KL from the truth to the updated predictive at the test inputs, averaged over the batch's labellings under the truth, and a decision's cost is the loss of \(\EPIG_H\)'s choice less EMI's.
Accuracy is the KL from the truth to the fitted predictive, averaged over the test inputs.

\paragraph{\Cref{fig:neural-testbed}.}
The local linear fits use a Gaussian kernel with bandwidth 0.16 in \(\log_{10} \Psi\) in (a), leaving out environments with \(\Psi \le 0\), and 0.06 nats in (b).
The percentiles in (a) are kernel-weighted and the standard errors in (b) approximate.
Markers are means over all environments in (a), unlike the medians of \Cref{tab:testbed-results}, and over those where the objectives disagree in (b).

\paragraph{Results.}
The per-environment Spearman correlation between the eight models' \(\Psi\) and distortion has mean 0.73 and median 0.79 over environments, and is positive in 996 of 1000.
\(\EPIG_H\) has a higher mean loss than EMI for every approximate model, by over two standard errors except for the incTNP (\(0.0009 \pm 0.0005\)).
Yet \(\EPIG_H\) chooses better in 37--49\% of disagreements.
Disagreement depends on the margin between EMI's two best candidates relative to the distortion.
Pooled over models, its rate falls from 0.74 in the lowest quintile of this ratio to 0.06 in the highest.
Over all models' disagreements, the cost of a disagreement has rank correlation 0.01 with \(\Psi\) and 0.21 with KL.
Across the eight models, its mean has rank correlation 0.24 with \(\Psi\) and 0.90 with the KL of \Cref{tab:testbed-results}.

\begin{table}[t]
\centering
\caption{Results on the neural testbed.
\(\Psi\) and \(\|\Delta\|\) are medians over environments.
Disagreement is the share of decisions where each form of EPIG differs from EMI.
Cost is averaged over all decisions, with agreements counting as zero, so the cost of a disagreement is Cost divided by the \(\EPIG_H\) disagreement.
KL measures accuracy before the update.
Both are in nats with one standard error.}
\label{tab:testbed-results}
\footnotesize
\setlength{\tabcolsep}{4pt}
\begin{tabular}{@{}l r r r r l l@{}}
\toprule
& & & \multicolumn{2}{c}{Disagreement} & & \\
\cmidrule(lr){4-5}
Model & \(\Psi\) & \(\|\Delta\|\) & \(\EPIG_H\) & \(\EPIG_{\KL}\) & Cost & KL \\
\midrule
Exact Bayes & \(0\) & \(0\) & \(0.00\) & \(0.00\) & \(0\) & \(0.171 \pm 0.003\) \\
IVON & \(0.0058\) & \(0.041\) & \(0.28\) & \(0.01\) & \(0.0084 \pm 0.0018\) & \(0.366 \pm 0.007\) \\
Laplace, KFAC & \(0.0067\) & \(0.059\) & \(0.40\) & \(0.01\) & \(0.0101 \pm 0.0020\) & \(0.310 \pm 0.007\) \\
Bayes by backprop & \(0.0102\) & \(0.094\) & \(0.80\) & \(0.10\) & \(0.0245 \pm 0.0027\) & \(0.362 \pm 0.007\) \\
incTNP & \(0.0103\) & \(0.097\) & \(0.26\) & \(0.02\) & \(0.0009 \pm 0.0005\) & \(0.168 \pm 0.003\) \\
Ensemble & \(0.0181\) & \(0.118\) & \(0.52\) & \(0.08\) & \(0.0195 \pm 0.0033\) & \(0.406 \pm 0.009\) \\
Laplace, full & \(0.0191\) & \(0.171\) & \(0.33\) & \(0.04\) & \(0.0071 \pm 0.0019\) & \(0.270 \pm 0.004\) \\
Laplace, diagonal & \(0.0474\) & \(0.318\) & \(0.38\) & \(0.08\) & \(0.0104 \pm 0.0020\) & \(0.269 \pm 0.004\) \\
MC dropout & \(0.117\) & \(0.425\) & \(0.54\) & \(0.20\) & \(0.0300 \pm 0.0069\) & \(0.478 \pm 0.012\) \\
\bottomrule
\end{tabular}
\end{table}

\end{document}